\documentclass[preprint,3p,onecolumn,authoryear]{elsarticle}
\usepackage{amssymb}
\usepackage{amsmath}
\journal{Applied Soft Computing}

\usepackage{float}
\usepackage{xcolor}
\usepackage{xspace}
\usepackage{booktabs}
\usepackage{makecell}
\usepackage{multirow}
\usepackage{verbatim}
\usepackage{subcaption}
\usepackage[colorlinks=true,linkcolor=blue]{hyperref}
\usepackage[linesnumbered,ruled,vlined]{algorithm2e}
\usepackage{enumerate}
\usepackage[shortlabels]{enumitem}

\usepackage{helvet}
\usepackage{etoolbox}
\usepackage{graphicx}
\usepackage{titlesec}
\usepackage{caption}
\usepackage{scrextend}
\usepackage{amsfonts}
\usepackage{bm}
\usepackage{amsmath}
\usepackage{soul}

\def\proposal{MO-EvoPruneDeepTL\xspace}
\def\Proposal{Multi-Objective Evolutionary Pruning for Deep Transfer Learning\xspace}

\begin{document}
\let\WriteBookmarks\relax
\def\floatpagepagefraction{1}
\def\textpagefraction{.001}

 \newenvironment{idea}{}{}
 
\begin{frontmatter}
\title{Multiobjective Evolutionary Pruning of Deep Neural Networks with Transfer Learning for improving their Performance and Robustness}



%

\author[1]{Javier Poyatos}
\ead{jpoyatosamador@ugr.es}

\author[1]{Daniel Molina\corref{mycorrespondingauthor}}
\cortext[mycorrespondingauthor]{Corresponding author}
\ead{dmolina@decsai.ugr.es}

\author[2]{Aitor Martínez-Seras}
\ead{aitor.martinez@tecnalia.com}


\author[2,3]{Javier Del Ser}
\ead{javier.delser@tecnalia.com}

\author[1]{Francisco Herrera}
\ead{herrera@decsai.ugr.es}

\affiliation[1]{organization={Department of Computer Science and Artificial Intelligence, Andalusian Research Institute in Data Science and Computational Intelligence (DaSCI), University of Granada},
    city={Granada},
    postcode={18071},
    country={Spain}}

\affiliation[2]{organization={TECNALIA, Basque Research \& Technology Alliance (BRTA)},
    city={Derio},
    postcode={48160},
    country={Spain}}

\affiliation[3]{organization={University of the Basque Country (UPV/EHU)},
    city={Bilbao},
    postcode={48013},
    country={Spain}}



\begin{abstract}
    Evolutionary Computation algorithms have been used to solve optimization problems in relation with architectural, hyper-parameter or training configuration, forging the field known today as Neural Architecture Search. These algorithms have been combined with other techniques such as the pruning of Neural Networks, which reduces the complexity of the network, and the Transfer Learning, which lets the import of knowledge from another problem related to the one at hand. The usage of several criteria to evaluate the quality of the evolutionary proposals is also a common case, in which the performance and complexity of the network are the most used criteria. This work proposes \proposal, a multi-objective evolutionary pruning algorithm. \proposal uses Transfer Learning to adapt the last layers of Deep Neural Networks, by replacing them with sparse layers evolved by a genetic algorithm, which guides the evolution based in the performance, complexity and robustness of the network, being the robustness a great quality indicator for the evolved models. We carry out different experiments with several datasets to assess the benefits of our proposal. Results show that our proposal achieves promising results in all the objectives, and direct relation are presented among them. The experiments also show that the most influential neurons help us explain which parts of the input images are the most relevant for the prediction of the pruned neural network. Lastly, by virtue of the diversity within the Pareto front of pruning patterns produced by the proposal, it is shown that an ensemble of differently pruned models improves the overall performance and robustness of the trained networks.
\end{abstract}



\end{frontmatter}

\newpage
\section{Introduction}\label{sec:introduction}

Evolutionary Computation (EC) refers to a family of global optimization algorithms inspired by biological evolution \citep{evocomp}. EC algorithms such as Evolutionary Algorithms (EA) \citep{EABook} have been used to solve several complex optimization problems which cannot be analytically solved in polynomial time. In many real-world optimization problems, there is not only one criterion or objective to improve, but several objectives to consider. Multi-Objective Evolutionary Algorithms (MOEAs) are an family of EAs capable of efficiently tackle optimization problems comprising several goals \citep{Deb2011}.

Structural search and, in particular, Neural Architecture Search (NAS), is one of the non-polynomial problems which has been approached with EAs over the years \citep{MARTINEZ2021161}. This problem consists of looking for neural network configurations that fit better one dataset by optimizing the performance or loss of the network in function of the selected evaluation metric \citep{Stanley200299}. There have been several Neural Networks (NN), and particularly when integrated with Deep Learning (DL) called as Deep Neural Network (DNN), to which NAS has been applied are well-known networks with one or more objectives \citep{pham2018efficient,supernet}.

Among other decision variables considered in NAS, this area has also approached the improvement of a NN by optimally pruning their neural connections. Pruning techniques seek to reduce the number of parameters of the network, targeting network architectures with less complexity. Usually this comes at the cost of a lower performance of the network. When a new learning task is present, a manner to compensate the lack of quality of data is the usage of the Transfer Learning (TL), whose most straightforward approximation is the usage of pre-trained network in very large datasets \citep{imagenet} for the extraction of features, followed by a specialization of the last layers of the network. Due to the fact that the number of trainable parameters of these last layers is lower, it is possible to avoid an early overfitting of the network, which can happen if there are few examples for large models. For those cases, the search of optimal pruning patters using evolutionary NAS is done for these last layers \citep{evoprunedeeptl}.


Robustness is one of the unavoidable requirements to ensure proper performance in scenarios where risks must be controlled and certain guarantees are needed to ensure the proper performance of the models \citep{iso1,iso2}. The combination of EAs together with techniques that allow evaluating the robustness of the models paves the way towards the creation of better models for all types of problems. It could be useful to incorporate robustness as a target, but unfortunately, robustness has been rarely considered an objective \citep{robustintro}. Robustness can be measured in several ways for a DNN model, one being the performance in Out-of-Distribution (OoD) detection problems \citep{hendrycks2016baseline}. This problem consists of detecting whether a new test instance queried to the model belongs to the distribution underneath the learning dataset i.e., the In-Distribution (InD) dataset or, instead, it belongs to another different distribution (correspondingly, the Out-Distribution dataset, OoD). 

The natural extension of NAS is the development of proposals with several objectives. In this scenario, the MOEAs can take place, as they evolve the networks meanwhile an optimization of several objectives is made \citep{lu2022neural,elsken2019neural}. The MONAS term arises as the union of MO algorithms which are used for NAS problems (MONAS). MONAS algorithms usually rely in several objectives, being a standard objective the performance of the network. The complexity of the network is a common second objective, which can be modeled as the number of parameters pruned form the network, network compression or other alternatives. More sophisticated proposals consider another objective based on the energy consumption or hardware device in use, among others \citep{nashardware,moodnas}. The addition of the robustness, with a OoD detection technique applied to the DL model being optimized, as an additional objective unleashes a new vision for the MONAS proposals.

The main hypothesis is the convenience of using a MOEA to evolve the pruning patterns of the fully-connected layers of a neural network via a sparse representation, simultaneously according to the generalization performance of the network, its complexity and the robustness of a OoD detection technique relying on the activation signals inside the network against samples that may or may not belong to the distribution of the training data.

\begin{idea}
This work finds its inspiration in the recent work in \citep{evoprunedeeptl}, in which dense layers are pruned using a configuration that define the active neurons. In the previous work, that configuration is evolved by using a binary genetic algorithm guided by the performance of the network. In this manuscript, the previous problem is reformulated to optimize the pruning patterns with a MOEA, in which the search is guided by the three previously mentioned objectives. Intuitively, a highly-pruned network may reduce its performance and the robustness of an OoD detection method that relies on the activations of the pruned network. For that reason, a minimum fraction of neurons must be active (i.e. non-pruned) to achieve balanced models with good balance (in the Pareto sense) between performance and robustness.

In this context, OoD detection falls within the umbrella of the Open-World Learning (OWL) paradigm \citep{owl,owlml}. OWL pursues models that are capable of learning in non-controlled environments, so that models become increasingly knowledgeable as they are queried with new data. However, OWL can also be considered one of the technologies supporting General Purpose Artificial Intelligence (GPAI), which is largely enabled by AI generating AI models \citep{aigenai,real2020automl}. Since this work proposes a MOEA to optimize DL models, it can be regarded as an example of AI enhancing AI.

In detail, this work proposes an approach based on the evolution of the pruning patterns of fully-connected layers using a MOEA, which we hereafter refer to as Multi-Objective Evolutionary Pruning for Deep Transfer Learning (\proposal). The goal of \proposal is to search for the best pruning patterns in the last layers of the NN to adapt them to the problem at hand. To accomplish this task, \proposal utilizes several techniques. To begin with, TL allows for the extraction of features by leveraging pretrained neural models, so that the specialization of the target NN takes place in the last fully-connected layers of the network hierarchy. At this point of the network pruning is as suitable mechanism to prune non-important features that do not contribute to the flow of information throughout the last part of the NN, which connects pretrained features to the output to be predicted. MOEA then emerges as an efficient method to solve the problem of finding good pruning patterns according to the aforementioned different objectives: performance, complexity and, robustness of the network. To measure the robustness of a model, an OoD detection technique is used, which is based on the capability of the model to detect unseen data in the training step. Ideally, robust models with good performance and low complexity should be desirable. However, the fact that pruning affects the activations throughout the last stage of the network causes that performance and robustness can be affected by the pruning intensity imposed by any given pruning pattern. This conflicting nature of the objectives under consideration is the rationale for seeking the optimal set of pruning patterns that best balance between them by using a MOEA. Finally, we will show that a byproduct of the estimated Pareto front is that NNs pruned by patterns belonging to the front can be combined together, yielding an ensemble model with increased performance and/or robustness with respect to any of its compounding NNs. This exposes that the pruning solutions give rise to NN models that present a sufficient diversity to improve their performance in accuracy and robustness over different value ranges of the objectives driving the search. 
\end{idea}

To assess the quality of \proposal, different experiments have been designed that allow inspecting several aspects of the performance of \proposal from different perspectives. To that end, the main purpose of the experimental setup is to provide an informed answer to the following research questions (RQ):
\begin{enumerate}[leftmargin=1.2cm, start=1,label={(\bfseries RQ\arabic*)}]
    \item How are the approximated Pareto fronts produced by the proposal in each of the considered datasets?
    \item Is there any remarkable pruning pattern that appears in all the solutions of the Pareto front?
    \item Do our models achieve an overall improvement in performance when combined through ensemble modeling?
\end{enumerate}

A general insight about these experiments is the the achievement of optimized networks in these objectives, but also that the evolutionary process gives rise to pruning patterns that maintain relevant neurons with information about the input of the model, and leads to the use of ensembles to further improve modeling performance in terms of generalization and robustness to OoD.


The rest of the article is structured as follows: Section~\ref{sec:relatedwork} briefly overviews background literature related to the proposal. Section~\ref{sec:proposal} shows the details of the proposed \proposal model. Section~\ref{sec:framework} presents the experimental framework designed to thoroughly examine the behavior of \proposal with respect to the RQ formulated above. In Section~\ref{sec:results}, we show and discuss in depth the results obtained by \proposal in the different experiments. Several indicators are presented to show the quality of \proposal. Finally, Section~\ref{sec:conclusions} draws the main conclusions from this study, as well as future research lines stimulated by our findings.

\section{Related work}\label{sec:relatedwork}

The aim of this section is to make a review of contributions to the literature about the key elements of this study: Neural Architecture Search (Subsection \ref{sec:nas}), Transfer Learning and Pruning of Convolutional Neural Networks (CNN, Subsection \ref{sec:tl-cnn}) and OoD detection (Subsection \ref{sec:ood}). The last paragraph of this section resumes the benefits of \proposal.

\subsection{Neural architecture search} \label{sec:nas}

The design of the NN that best fits for the problem at hand is a challenging task. The search for the best design of the network is also considered as another problem, as it is necessary to find the best architecture that optimally fits the data. In this context, NAS has achieved a great importance in this area. The main purpose of NAS proposals is the search for the best design of the NN to solve the considered problem.

First NAS-based proposals started to emerge in the beginning of this century. NEAT, presented in \citep{Stanley200299} was a pioneering proposal about how EAs --- specifically, a Genetic Algorithm (GA) -- can be used to evolve NNs. They showed that a constructive modeling of the NN with the benefits of the GA can lead to optimized NN topologies. The natural extension of this seminal work allowing for the evolution of DNN was presented years after in \citep{miikkulainen2019evolving}, in which the authors use a co-evolutionary algorithm based on the co-operation scheme to evolve DNN.

In the last years, more NAS proposals have been developed. One of them is EvoDeep \citep{Comput2018}, in which the authors create an EA with specific operators to create and evolve DL models from scratch. More examples of the importance of NAS come with the next proposals. In \citep{dufourq}, authors propose another EA to perform the evolution of NN, similarly to the previous proposal, but with a difference in relation to the fitness function, which is influenced by the accuracy and complexity of the network. The other example is presented in \citep{assunccao2019denser}. In this case, the evolution comes in two different ways: topology and parameters of the Convolutional Neural Networks (CNN). In \citep{TRIVEDI2018525}, the authors propose a GA that evolves the weights of the softmax layer to improve the performance of the NN. Suganuma et al. propose in \citep{suganuma} a $(1+ \lambda)$ evolutionary strategy to evolve DNN. In 2020, \citep{real2020automl} presents an advanced technique that automatically searches for the best model, operating from scratch and obtaining a good performance with the problems at hand. The use of NAS has been applied in other areas like the Reinforcement Learning (RL). In that area, there is a great example of NAS \citep{nasrl}. In that work, authors use a recurrent network (RNN) to generate the model descriptions of NN and train this RNN with RL to maximize the expected accuracy of the generated architectures on a validation set.

There are more examples of NAS in the literature like the NAS algorithm which comprises of two surrogates through a supernet, with the objective of improving the gradient descent training efficiency \citep{nsganetv2}. Another NAS comes in \citep{surrogate}, in which the authors propose a pipeline with also a surrogate NAS applied to real-time semantic segmentation. They manage to convert the original NAS task into an ordinary MO optimization problem.

Lastly, there are more advanced techniques of NAS and EA given by \citep{real2019regularized}, in which a new model for evolving a classifier is presented, and by \citep{real2020automl}, in which the authors propose AutoML-Zero, an evolutionary search to build a model from scratch (with low-level primitives for feature combination and neuron training) which is able to get a great performance over the addressed problem.

The main characteristic of the previous NAS proposals is the evolution of the DL model guided by a single objective, usually the accuracy or another that measures the performance of the network. The following proposals share a common aspect: the evolution of the model is done using more than one objective. This leads to the algorithms in the field of MONAS.

We can find several approaches of MONAS that have been applied to diverse fields with great results. One of them is presented in \citep{Elsken2019EfficientMN}. This work proposes a MONAS that lets the approximation of the Pareto-front of all the architectures. In relation to medical images area, in \citep{BALDEONCALISTO2020325} a MONAS that evolves both accuracy and model size is proposed. Moreover, following this research in medical images, in \citep{baldeon}, the authors use a MO evolutionary based algorithm that minimizes both the expected segmentation error and number of parameters in the network. Another interesting work is presented in \citep{calistonas}, in which they have created a pipeline for the automatic design of neural architectures while optimizing the network's accuracy and size.

Typically, MONAS evaluate two or three objectives. A common objective is usually the performance of the network. The others objectives are related with the complexity of the network and other empirical and measurable objectives. In \citep{moevo1}, the authors propose a MOEA for the design of DNN for image classification, adopting the classification performance and the number of floating-point operations as its objectives. Another example is DeepMaker, \citep{LONI2020102989}, which is a MOEA approach that considers both the accuracy of the network and its size to evolve robust DNN architectures for embedded devices.

There are some well-known MOEAs in the literature. One of them is NSGA-II. A new version of it has been developed to use it for NAS \citep{nsganet}, called NSGA-Net. This proposal looks for the best architecture through a three-step search based on an initialization step, followed by an exploration step that performs the EA operators to create new architectures, and an exploitation step that uses the previous knowledge of all the evaluated architectures.

\subsection{Transfer learning and pruning} \label{sec:tl-cnn}

One of the objectives that EAs used for NAS usually aim to optimize is the complexity of the network. NN are structures with a great amount of parameters. These networks are composed of two main parts. The first one extracts the main features of the problem, i.e., learns to distinguish the patterns of the images (when working with image classification) and the second part is responsible to classify these patterns into several classes.

In this context, TL appears as a figure that helps in the learning process when there are few data, i.e., prevents the overfitting when the input examples is not large \citep{tl}. TL is a DL mechanism encompassing a broad family of techniques. The most common method of TL with DL is the usage of a previous network structure with pre-trained parameters in a similar problem to the related task, being trained with huge datasets like \citep{imagenet}. This fact involves the usage of a DL model with fixed and pre-trained weights in the convolutional layers with a dataset and then add and train several layers to adapt the network to a different classification problem \citep{khan2019novel}.

Another technique to reduce the complexity of the networks is pruning. Pruning a CNN model consists of reducing the parameters of the model, but it may lead into a decrease of the performance of the model.  
Several approaches to prune networks have been developed over the years, such as \citep{han2015learning,srinivas2015data}. These methods have been already used in several problems, rendering great performance.

An example of the fusion of EAs, DNN and pruning is shown in \citep{WANG2020247}, which proposes a novel approach based on a combination of pruning CNN of sparse layers (layer with fewer connections between neurons) guided by a GA. The main consequence of this study is the reduction of a great fraction of the network, but at the penalty of a lower generalization performance of the network.

Following the idea of EAs and DNN, in \citep{evoprunedeeptl} the authors propose also propose a combination of sparse layers and a GA. They have shown that pruning can be done in a TL scheme with sparse layers and EAs. Their proposal is only guided by the performance of the model, but they also achieve a great reduction in the optimized sparse layers. 

\subsection{Out of distribution detection}\label{sec:ood}

Robustness is a term that has been used with related yet different meanings among the literature of the Machine Learning (ML) community. In this work, we refer to the model's ability to handle the unknown, to detect whether it has been queried with an example of a not learned distribution, therefore refusing to make the classification it has been trained to do. This is precisely what the OoD detection framework measures.

In this problem, a model learns to classify instances in the different classes from a training dataset that is sampled from a distribution, namely the InD. After the process, the model is asked to correctly distinguish between test examples that are drawn equally from either the  InD  or from a semantically different distribution, the OoD dataset \citep{yang2021generalized}. The term semantically different refers to the fact that the classes contained in this foreign distribution are distinct from the ones present in the InD. As ML and DNN model are not natively prepared for this task, an OoD detection technique is wrapped around the model to allow this behavior. Typically, these techniques are based on creating a score for every example processed by the model, such that the score obtained by an OoD instance is significantly different from the one obtained by a InD example. Then, by simply defining a threshold on this score, the model can decide whether an instance is from the in or out distribution.

A great variety of methods exist in the literature, which was started by \cite{hendrycks2016baseline}, where the so-called \emph{baseline} method was introduced. It relies on the simple observation that InD instances tend to have greater Maximum Softmax Probability, the softmax probability of the predicted class. By simply applying a threshold to this score, they achieved acceptable performance on many classification problems. In \cite{liang2017odin}, this idea was refined by applying temperature scaling to the softmax probabilities, what further separates apart from each other the distributions of the scores of the in- and out- distribution samples probabilities. Authors also implemented an input preprocessing pipeline that enhanced a bit the performance by adding a small quantity of gradient and softmax dependent noise. The paper presented in \cite{lee2018simple}, instead of using the softmax probabilities, exploits the feature space of the layer right before softmax and assumes that it follows a multivariate Gaussian distribution, enabling the calculation of its mean and variance for every sample. After creating a class-conditional distribution utilizing the training samples, the score for every test sample is the closest Mahalanobis distance between the sample and the calculated Gaussian class-conditional distributions.

The technique proposed in \citep{hendrycks2018deep}, in contrast to previous works, focuses on modifying model's training by adding a term to the loss function (that depends on the classification of the task, density estimation, etc.), helping the model learn heuristics that will improve the performance of other OoD methods applied afterwards. This new term needs to be trained with OoD data, which can be obtained by leveraging the large amount of data publicly available on the internet. Authors prove that the learned heuristics for arbitrary OoD datasets generalize well to other unseen OoD data. Thereafter, \citep{liu2020energy} based its detector in what they called the free energy function, that combines concepts of the energy-based models with modern neural networks and their capability of assigning a scalar to every instance fed to the model without changing its parametrization. Specifically, the free energy function is based on the logits of the network, and the work empirically demonstrates that OoD instances tend to have higher energy, enabling the distinction between InD and OoD data. In the following work in \citep{lin2021mood} exploited the idea that easy OoD samples can be detected by leveraging low-level statistics. On this basis, several intermediate classifiers are trained at different depths and each example is outputted through one of them depending on its complexity. To measure complexity, a function based on the number of bits used to encode the compressed image is harnessed. The OoD scoring function employed is the above presented energy function adapted to the corresponding depth.

Although only a few research contributions are presented in this work, it must be noted that the OoD problem has been widely studied in the literature \citep{salehi2021unified}, with proposals ranging from the more complex and well performing ones to the more simple yet effective ones. As the aim of this paper is to 
show that the robustness in the OoD can be affected when the pruning of the network is done. Therefore, the OoD detection method will be selected to be computationally cheap yet effective, to not add computational complexity to the MOEA.

In this section, a review of the related work from three different perspectives has been presented. Terms like MONAS, MOEA are important as this work presents a new work about these topics. Moreover, it is based on a TL scheme in which an evolutionary pruning of the last layers is done. In the last years, several proposals have been published over these topics, i.e, MOEAs that search for the best architecture attending to one or more objectives and also pruning approaches for CNNs. However, this study introduces a new manner to guide the evolutionary pruning of the models with the usage of a OoD mechanism. \proposal tries to solve the problem of achieving robust models with high performance and least active neurons. This scheme, a MOEA that performs pruning in the last layers (TL paradigm) with three objectives is a new contribution to all this fields at the same time.

\section{\Proposal} \label{sec:proposal}

This section is devised to explain the details of \proposal. First, in Subsection \ref{sec:pf} the formulation to the problem at hand is presented. In Subsection \ref{sec:objectives}, we will describe the objectives used to guide the search. Then, in Subsection \ref{sec:pood}, the description of the OoD detector is explained. The DL and network schemes are shown in Subsection \ref{sec:pdl}. Finally, in Subsection \ref{sec:pevo}, the evolutionary parts of \proposal are described.

\subsection{Problem formulation}\label{sec:pf}

This section aims at defining and explaining the mathematical components that circumscribe \proposal. We explore different concepts needed to fully understand the basics of our study.

We define the concept of dataset. Mathematically, we define a training dataset $\mathcal{D}\doteq \{(\mathbf{x}_i,y_i)\}_{i=1}^N$ composed by $N$ (instance, label) pairs. Such a dataset is split in training, validation and test partitions, such that $\mathcal{D}=\mathcal{D}_{tr}\cup\mathcal{D}_{val}\cup\mathcal{D}_{test}$ with $|\mathcal{D}_{tr}|=N_{tr}$.

Another important concept to keep in mind is the model. We define a model $M_{\theta}$ to represent the relationship between its input $\mathcal{X}$ and its corresponding output $y\in\{1,\ldots,Y\}$, where $Y$ denotes the number of classes present in $\mathcal{D}$. Learning the parameter values $\theta^\ast$ that best model this relationship can be accomplished by using a learning algorithm $\theta^\ast=\text{ALG}(M;\mathcal{D}_{tr})$ that aims to minimize a measure of the difference (\emph{loss}) between the model's output and its ground truth over the training instances in $\mathcal{D}_{tr}$ (e.g. gradient back-propagation in neural networks). In what follows $M_\theta$ is assumed to be a NN, so that $\theta$ represent the totality of trainable weights in its neural connections.

In the context of TL for classification tasks, the NN $M_\theta$ is assumed to be
composed by a pre-trained feature extractor $F_{\phi}(\mathbf{x})$ (whose parameters $\phi$ are kept fixed while $\text{ALG}(\cdot)$ operates), and a dense (i.e. fully-connected) part $G_\theta(\cdot)$ that maps the output of the feature extractor to the class/label to be predicted. Therefore, after tuning the trainable parameters of the network as $\theta^\ast = \text{ALG}(G; \mathcal{D}_{tr})$, the class $\widehat{y}\in\{1,\ldots,Y\}$ predicted for an input instance $\mathbf{x}$ is given by:
\begin{equation}
\widehat{y}=(F \circ G_{\theta^\ast})(\mathbf{x}) = G_{\theta^\ast}(F(\mathbf{x})),
\end{equation}
where $\circ$ denotes composition of functions. When predictions are issued over the validation partition $\mathcal{D}_{val}$, a measure of accuracy can be done by simply accounting for the ratio of correct predictions to the overall size of the set, i.e. $\text{ACC}_{val} = (1/N_{val})\cdot \sum_{i\in\mathcal{D}_{val}} \mathbb{I}(\widehat{y}_i=y_i)$, where $\mathbb{I}(\cdot)$ equals $1$ if its argument is true ($0$ otherwise).

Bearing this notation in mind, pruning can be defined as a binary vector $\mathbf{p}=\{p_j\}_{j=1}^P$, where $P$ denotes the length of the feature vectors extracted by $F_\phi(\mathbf{x})$ for every input instance to the network. As such, $p_j=0$ indicates that neural links that connect the $j$-th component of the feature vector to the rest of neurons in the dense part $G_\theta(\cdot)$ of the network are disconnected, causing that all trainable parameters from this disconnected input to the output of the overall model are pruned. Conversely, if $p_j=1$ the $j$-th input neuron is connected to the densely connected layers of the neural hierarchy. By extending the previous notation, the training algorithm is redefined to $\theta^\ast(\mathbf{p})=\text{ALG}(G;\mathcal{D}_{tr},\mathbf{p})$ to account from the fact that the network has been pruned as per $\mathbf{p}$. This dependence of the trained parameters on the pruned vector propagate to the measure of accuracy over the validation instances, yielding $\text{ACC}_{val}(\mathbf{p})$. Likewise, a measure of the reduction of the number of trainable parameters can be also computed for a given pruning vector $\mathbf{p}$ relative to the case when no pruning is performed (i.e., $\mathbf{p}=\mathbf{1}\doteq\{1\}_{j=1}^P$) as $\text{MEM}(\mathbf{p})=|\theta(\mathbf{p})|/|\theta(\mathbf{1})|$.

Intuitively, a good pruning strategy should consider the balance between the reduced number of trainable parameters and its impact on the accuracy when addressing the modeling task at hand. Reducing the amount of parameters to be stored has practical benefits in terms of memory space, and can yield a lower inference latency when the trained model is queried.

A third dimension of the network that can be affected by pruning is its capacity to detect  OoD  instances. A significant fraction of the techniques proposed so far for identifying query samples that deviate from the distribution of training data rely on the network dynamics between neurons while the instance flows through the network. This is the case of ODIN \citep{liang2017odin}, BASELINE \citep{hendrycks2016baseline} and ENERGY \citep{liu2020energy}, among others. To quantify the capability of a pruned network $M_{\theta^\ast(\mathbf{p})}$ to detect OoD instances, we utilize other datasets $\smash{\bm{\mathcal{D}}^\prime = \{\mathcal{D}_d^\prime\}_{d=1}^{D_{OoD}}}$ different from $\mathcal{D}$, whose instances $(\mathbf{x}^\prime,y^\prime)$ are assumed to be representative of the OoD test instances with which the model can be queried in practice. An OoD detection technique $T_{OoD}(\mathbf{x};M_{\theta^\ast(\mathbf{p})})\equiv T_{OoD}(\mathbf{x})$ processes the activations triggered by $\mathbf{x}$ throughout the trained pruned model $M_{\theta^\ast(\mathbf{p})}$ so as to decide whether $\mathbf{x}$ is an  InD  ($T_{OoD}(\mathbf{x})=0$) or an OoD instance (corr. $T_{OoD}(\mathbf{x})=1$). This being said, true positive and false negative rates can be computed for $T(\mathbf{x})$ over the test subset $\mathcal{D}_{test}$ of $\mathcal{D}$ and random $N_{val}/D_{OoD}$-sized samples drawn from every other dataset $\mathcal{D}_d^\prime$, which can be aggregated in a compound performance metric. Among other choices for this purpose, we consider the AUROC measure $\text{AUROC}(\mathbf{p})$, which measures the ability of $T(\cdot)$ to discriminate between positive and negative examples. This measure is set dependent on $\mathbf{p}$ in accordance with previous notation, as $T(\mathbf{x})$ operates on the neural activations stimulated by $\mathbf{x}$.

\subsection{Objectives of \proposal} \label{sec:objectives}

This section introduces the objectives that guide \proposal during its evolutionary process. We define them using the notation previously commented in Subsection \ref{sec:pf}.

The optimization problem addressed in this work aims to discover the set of Pareto-optimal pruning vectors $\{\mathbf{p}_k^{opt}\}_{k=1}^K$ that best balance between three objectives:

\begin{enumerate}
    \item The modeling performance of the pruned model over dataset $\mathcal{D}$. This performance is measured with the accuracy over the test dataset ($\mathcal{D}_{test}$). It is the percentage of well classified images out of the total set of images.
    \item The number of active neurons left after the pruning operation. The number of active neurons corresponds with the remaining active connections after the pruning and evolutionary process.
    \item The capability of an OoD detection technique to discriminate between  OoD and InD
     data by inspecting the activations inside the pruned model.
\end{enumerate}

Mathematically:
\begin{align}
\{\mathbf{p}_k^{opt}\}_{k=1}^K &=\arg_{\mathbf{p}\in\{0,1\}^P} \left[\max \text{ACC}_{val}(\mathbf{p}),\min \text{MEM}(\mathbf{p}), \max \text{AUROC}(\mathbf{p})\right],\\
\textrm{s.t.} \quad & \mathcal{D}: \textrm{In-distribution dataset}, \\
& \mathcal{D}_1^\prime, \ldots, \mathcal{D}_{D_{OoD}}^\prime: \textrm{Out-of-distribution datasets}, \\
& F_{\phi}(\mathbf{x}): \textrm{Pre-trained feature extractor},\\
& T(\mathbf{x}): \textrm{Out-of-distribution detection technique}.
\end{align}

\subsection{Out of Distribution detector of \proposal} \label{sec:pood}

In the following subsection the technique selected to assess the OoD performance of the pruned models is presented, along with a clarification about the metrics used to measure it.

Due to the fact that every new child of the population in the evolutionary algorithm must have its OoD performance correctly assessed, the chose method should not entail a big computational burden while maintaining a sufficient effectiveness in detecting OoD samples. The technique presented in \citep{liang2018enhancing}, ODIN, fulfills these requirements and is the selected one.

Before explaining ODIN, the already mentioned performance metric used in this study must be clarified, namely the AUROC or \emph{Area Under the Receiver Operation Characteristic curve}. It is a threshold-independent metric for binary classification that can be considered as the probability that the model ranks a random positive example with higher score than a random negative example. Is defined as $\text{TPR}/\text{FPR}$, which stand for True Positive Rate and False Positive Rate respectively and can be computed as $\text{TPR} = \text{TP}/(\text{TP}+\text{FN})$ and $\text{FPR}=\text{FP}/(\text{FP}+\text{TN})$. Therefore, in order to compute the AUROC, the FPR value for every TPR needs to be calculated. In this work, TP is used to refer to an in-distribution sample correctly classified as such, whereas a TN represents an OoD sample detected correctly by the OoD detector.

The basic principle of ODIN is to use maximum softmax probability with temperature scaling as the OoD score for every sample, defined by the expression
\begin{equation}\label{eq:fodin}
f_{ODIN}(\mathbf{x};T) = \mathrm{max}_i(S_i(\mathbf{x};T)) = S_{\hat{y}}(\mathbf{x};T),
\end{equation}

where $S_i(x;T)$ is the softmax probability of the $i^th$ class for the input instance $\mathbf{x}$, scaled by a temperature parameter $T \in \mathbb{R}^+$. This scaled softmax can be calculated as:
\begin{equation}\label{eq:softmaxTemp}
S_i(x;T) = \frac{\mathrm{exp}(h_i(\mathbf{x})/T)}{ \sum^{N}_{j=1} \mathrm{exp}(h_i(\mathbf{x})/T)}.
\end{equation}
where $h_i(\mathbf{x}) \ | \ i \in \{1,\ldots,Y\}$ are the logits, the values prior to the softmax activation function. Then, and in accordance with notation presented in Subsection \ref{sec:pf}, the OoD detection technique $T_{OoD}$, ODIN in this case $T_{ODIN}$, will output a 1 if the instance's score is below a defined threshold, indicating that is considered an out-of-distribution sample, outputting a 0 otherwise:
\begin{equation}\label{eq:OoDdecision}
    \mathbf{x}\text{ belongs to}\begin{cases}
      \text{in-distribution} & \text{if } T_{ODIN}(\mathbf{x};T;\lambda) = 0 \iff f_{ODIN}(\mathbf{x};T) \geq \lambda, \\
      \text{out-distribution} & \text{if } T_{ODIN}(\mathbf{x};T;\lambda) = 1 \iff f_{ODIN}(\mathbf{x};T) < \lambda.
    \end{cases}
\end{equation}

It is important to remark that ODIN also uses an input preprocessing pipeline to further improve its performance in the OoD detection problem, but that in this study it will be discarded for the sake of reducing the computational burden of the algorithm.

So, in order to implement ODIN, the below presented steps must be followed. First, the model $M_\theta$ must be trained using the training set $\mathcal{D}_{tr}$ of the in-distribution dataset $\mathcal{D}$. Then the logits of the instances of the test set $\mathcal{D}_{test}$ must be extracted for the sake of calculating the temperature scaled softmax outputs using the equation \eqref{eq:softmaxTemp}. The OoD score of each input instance $f_{ODIN}(\mathbf{x};T)$ will be the maximum of these scaled softmax outputs, i.e., the value corresponding to the predicted class, as expression \eqref{eq:fodin} indicates.

Next, same operation must be repeated with the the out-of-distribution detection set, composed by samples drawn from every other dataset $\mathcal{D}_d^\prime$ as indicated in Subsection \ref{sec:pf}. In this manner, we have created two distributions of OoD scores: one for the in-distribution samples of $\mathcal{D}_{test}$ and other for the out-of-distribution ones. Now, the threshold on the score for each TPR is defined by using the score distribution of test instances and equation \eqref{eq:OoDdecision}. The corresponding FPR for each TPR is computed by employing the OoD distribution and the defined threshold, obtaining a set of [TPR, FPR] values that compose the ROC curve. Finally, from this curve the AUROC can be computed, therefore obtaining the desired robustness score for the model $M_\theta$ and in-distribution dataset $\mathcal{D}$.

In this study, in order to evaluate the robustness of each model, a practical approach is used, which involves the usage of the OoD detector with the other datasets that are not covered in training phase. However, the design of the algorithm accommodates any other dataset as an OoD evaluation dataset.

\subsection{Network characteristics of \proposal} \label{sec:pdl}

In this subsection, we introduce the characteristics of the used network in \proposal. In our study, we use the TL paradigm, i.e., the weights of the convolutional phase are imported and fixed from another trained network in a similar task. For that reason, the DL model we use works as a feature extractor. The images pass through the network and it extracts their main features. These features correspond with the neurons which are evolved by the evolutionary components of \proposal. 

\begin{figure}[htp]
        \centering
        \includegraphics[width=0.9\textwidth]{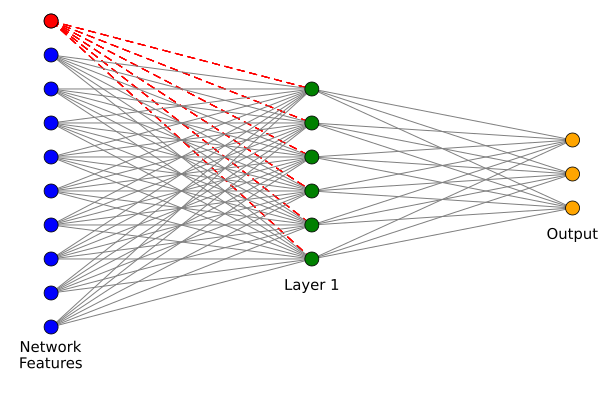}
        \caption{Pruning method of MO-EvoPruneDeepTL.}
        \label{fig:pruningmethod}
    \end{figure}

    The chosen network for this study is ResNet-50. The output of this network is a vector of 2048 features or characteristics. They are used as the input for the last layers of the neural network. In our case, following the research in \citep{evoprunedeeptl}, the last part of the neural network is composed by an input layer that receives an input vector of 2048 features (i.e., the output of the ResNet-50), followed by a hidden layer of 512 neurons and, finally, an output layer with as many neurons as classes defined in the problem at hand. This architecture is depicted in Figure \ref{fig:pruningmethod}, wherein \emph{Layer 1} corresponds to our intermediate layer of 512 neurons, and \emph{Network Features} denote the vector of 2048 features extracted from ResNet-50. We highlight in red the connections affected when a neuron is pruned. More specifically, each feature contributes to the neurons of the intermediate layer. The solver learns to distinguish which features are the most important and which are not, so that the whole set of connections from irrelevant features onward is eliminated, thereby not contributing to the rest of the intermediate layer.
    
    Following the previous idea, this network has fewer connections than a standard fully-connected layer, in which each neuron is connected to the next group of neurons. This type of layer is referred to as \emph{sparse layer}. The genetic algorithm is in charge of finding an optimal pruning pattern for that sparse layer, in which the chromosome representing the pruning pattern can be decoded to an adjacency matrix. Figure \ref{fig:pruningmethod} shows that this matrix defines the structural composition (connections) of the layer in the neural network. In particular, binary entries in the adjacency matrix correspond to the connections between the blue and green circles, so that certain connections will be removed (red lines) as a result of the pruning operation, rendering the sparse nature of the matrix and the layer itself.

\subsection{Evolutionary components of \proposal} \label{sec:pevo}

In this section, we introduce the evolutionary components of \proposal. It is a MOEA, called Non-Dominated Sorting Genetic Algorithm II (NSGA-II) \citep{nsga}. The population of networks is evolved using the common operators from this GA, but, in this case, only two individuals are used for the evolution as parents. As a result, two offspring individuals are produced. Our \proposal uses a binary encoding strategy, which represents if a neuron is active or not. A neuron is active if its gen is 1 or not active if it is 0. Thanks to this direct encoding approach, each gen determines uniquely a neuron in the decoded network.

The initialization of the chromosomes correspond with a random discrete initialization in $[0,1]$, the selection is done using a binary tournament selection method, whereas the replacement strategy is the dual strategy of ranges of Pareto dominance and crowding distance of NSGA-II. Finally, the crossover and mutation operator are outlined:


\textbf{Crossover}: the crossover operator used in \proposal is the uniform crossover. This operator defines two new individuals from two parents. Mathematically, given these two parents \textbf{p} and \textbf{q}, where $\textbf{p}=\{p_i\}_{i=1}^P$ and $\textbf{q}=\{q_i\}_{i=1}^P$ and their length is $P$, the resultant offsprings $\textbf{p'}=\{p'_i\}_{i=1}^P$ and $\textbf{q'}=\{q'_i\}_{i=1}^P$ (also with length $P$) are generated using these equations:

\begin{equation}\label{crossover_operator}
  \begin{split}
        p_i' = \left\{
            \begin{array}{ll}
            p_i & \mbox{if } r \le 0.5\\
            q_i & \mbox{otherwise}\\
            \end{array}
        \right. \\
        q_i' = \left\{
            \begin{array}{ll}
            q_i & \mbox{if } r \le 0.5\\
            p_i & \mbox{otherwise}\\
            \end{array}
        \right.
  \end{split}
\end{equation}
where $r$ is the realization of a continuous random variable with support over the range $[0.0, 1.0]$. This operator creates two individuals using information of the genes of both parents. Each position $i$ of the new individual takes the value of the gene of $\textbf{p}$ or $\textbf{q}$ until the offspring is fully created.

\textbf{Mutation}: the mutation performed by \proposal is the Bit Flip mutation. This operator needs a mutation probability defined by $mut_p$. Thus, for each chromosome, all of its genes can be mutated if the mutation is really performed, which means changing the value of the gene from active to not active or vice versa. The parameter that controls if a gene is flipped or not is $mut_p$.

Next, we give a brief explanation about the process that \proposal performs. First, we need to know the data required by \proposal, which is the dataset for training the network and its test dataset, the InD data, and also the OoD data, so that each model can also be tested on it. Lastly, the configuration of the GA and of the network are also required.

Once all the data is gathered, then the evolutionary process takes place. Algorithm \ref{alg:proposal} shows the pseudocode of \proposal. The beginning of the process is the standard procedure of initialization and evaluation of the initial population (lines 1 and 2). Then, the evolution is performed. In each generation, the operators are being executed sequentially. The parents are selected using the selection operator (line 4). After that, they generate their offspring using the crossover operator (line 5) which are mutated using the mutation operator (line 6). Then, both children are evaluated to obtain the values of the objectives that guide the evolutionary process. Thus, for each child, its chromosome is decoded into a sparse network (line 8) which is trained using the train set of the InD data (line 9). Then, the information contained in the logits is passed through the OoD detector which determines the robustness of the child using the AUROC metric (line 10). The accuracy is calculated using the test set of the InD (line 11) and the complexity of the network is also achieved using the number of neurons which are active in the child chromosome (line 12). Then, the objectives are retained as part of the information of the child for further generations (line 13).

\begin{algorithm}[!hbt]
\SetAlgoLined
 \caption{\proposal}
\label{alg:proposal}

\SetKwInOut{Input}{Input}
\SetKwInOut{Output}{Output}
\Input{InD dataset, OoD dataset, configuration of the GA and configuration of the network}
\Output{Evolved pruned network}
 Initialization of individuals of the population using the initialization operator\;
 Evaluation of the initial population (see lines 9-14)\;
 \While{evaluations $<$ max\_evals} {
  Parent selection using binary tournament\;
  Generate offsprings using uniform crossover\;
  Mutation of individuals using the bit flip mutation;

  \For{each child $p$ in children population} {
  
    SparseNetwork$\boldsymbol{_p}$  $\leftarrow$ Decodification of child chromosome\;
    SparseTrainedNetwork$\boldsymbol{_p}$ $\leftarrow$ Train SparseNetwork$\boldsymbol{_p}$ using the train set of InD data\;
    AurocChild$\boldsymbol{_p}$ $\leftarrow$ Robustness metric of child using OoD data\;
    AccChild$\boldsymbol{_p}$ $\leftarrow$ Accuracy of SparseTrainedNetwork$\boldsymbol{_p}$ evaluated in test set of InD data\;
    ComplexChild$\boldsymbol{_p}$ $\leftarrow$ Number of active neurons in SparseNetwork$\boldsymbol{_p}$\;
    SolutionVector(AccChild,ComplexChild,AurocChild)$\boldsymbol{_p}$\;
    evaluations+=1\;
  }

  Replacement Strategy\;
 }
\end{algorithm}

\section{Experimental framework} \label{sec:framework}

This section is intended to describe the framework surrounding the experiments conducted in this study. In Subsection \ref{sec:di}, a detailed description of the datasets is given. Then, Subsection \ref{sec:tns} shows the values of the parameters and the network setup of \proposal in the experiments.

\subsection{Dataset information} \label{sec:di}

In this study we have selected several datasets which fit in our working environment. These datasets represent a good choice for TL approaches due to their size, as the training and inference times are lower. Thus, these datasets are suitable for problems related with population metaheuristics, since a large number of individuals will be evaluated. We present a brief description of each dataset:

\begin{itemize}
    \item CATARACT \citep{Eyes} is a dataset related with the medical environment. It classifies different types of eye diseases.
    \item LEAVES \citep{RAUF2019104340} is a dataset that is composed of images of different types of leaves, since healthy to unhealthy with different shades of green.
    \item PAINTING is related to the painting environment \citep{Musemart}. This dataset is composed of images which represent different types of paintings.
    \item PLANTS is dataset which presents a great variety of leaves and plants, which ranges from tomato, or corn plants to other leaves, among others \citep{plantdoc}.
    \item RPS \citep{rps} is a dataset whose purpose is to distinguish the gesture of the hands in the popular Rock Paper Scissors game from artificially-created images with different positions and skin colors.
    \item SRSMAS is based on the marine world whose aim is to classify different coral reef types \citep{SRSMAS}.
\end{itemize}

\begin{figure}[!hbtp]
    \begin{center}
        \begin{minipage}[b]{.95\textwidth}
        \includegraphics[width=0.25\linewidth]{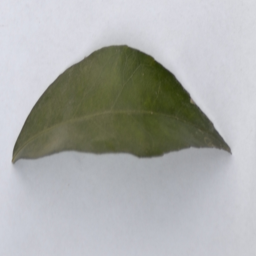}\qquad
        \includegraphics[width=0.25\linewidth]{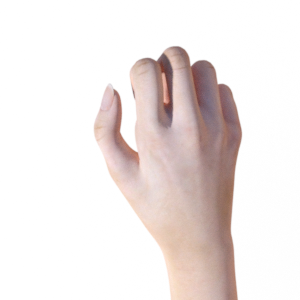} \qquad
        \includegraphics[width=0.32\linewidth]{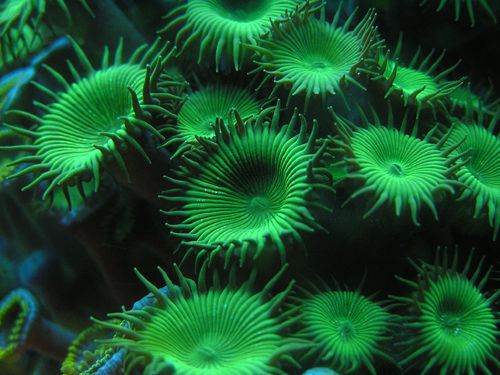}
        \end{minipage}

        \begin{minipage}[b]{.95\textwidth}
        \includegraphics[width=0.25\linewidth]{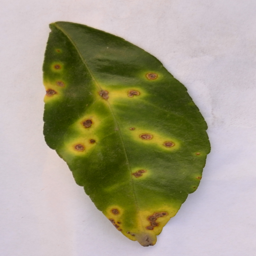}\qquad
        \includegraphics[width=0.25\linewidth]{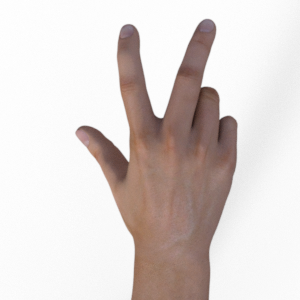} \qquad
        \includegraphics[width=0.32\linewidth]{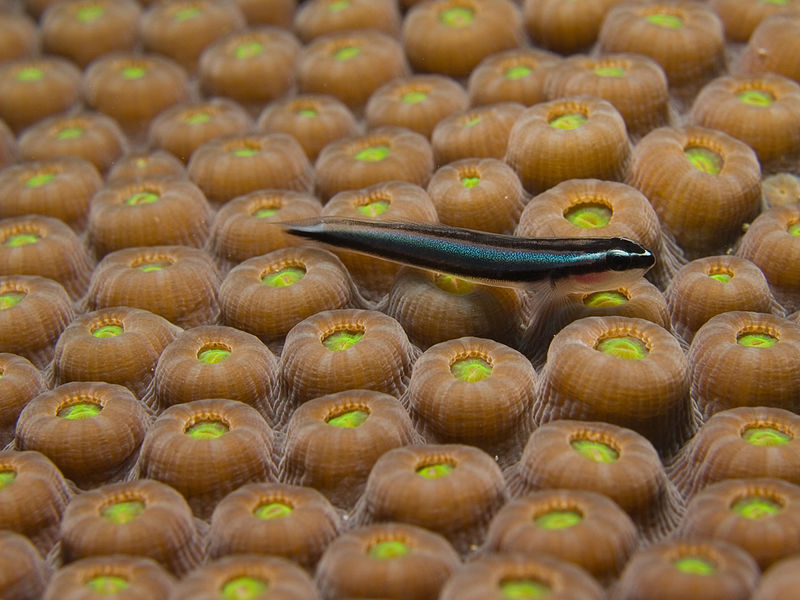}
        \end{minipage}


    \end{center}
\caption{Images of datasets. Left: LEAVES examples. Middle: RPS examples. Right: SRSMAS examples.}
\label{fig:images-datasets}
\end{figure}

Next, we show some examples for several of the above datasets are shown in Fig. \ref{fig:images-datasets}.

Finally, we highlight the main characteristics in quantitative terms of instances, classes and metrics with non-pruned networks for each dataset. Table \ref{tbl:datasets} show these numbers.

\begin{table}[!h]
\centering
\caption{Datasets used in the experiments.}
\label{tbl:datasets}
    \begin{tabular}{lccccc}
    \toprule
         \textbf{Dataset} & \textbf{Image Size} & \makecell{$L$\\(\textbf{\# classes})} & \makecell{\textbf{\# Instances} \\ (\textbf{train / test})} & \makecell{\textbf{Accuracy}\\(\textbf{No Pruning})}  & \makecell{\textbf{AUROC}\\(\textbf{No Pruning})} \\
        \midrule
        CATARACT      & $(256,256)$ & 4 & 480 / 121 & 0.732 &  0.870 \\
        LEAVES       & $(256,256)$ & 4 & 476 / 120 &  0.935 &  0.960 \\
        PAINTING     & $(256,256)$ & 5 & 7721 / 856 &  0.951 & 0.990 \\
        PLANTS       & $(100,100)$ & 27 & 2340 / 236 & 0.480 & 0.820 \\
        RPS          & $(300,300)$ & 3 & 2520 / 372 & 0.954 & 0.934 \\
        SRSMAS        & $(299,299)$ & 14 & 333 / 76 & 0.885 & 0.999 \\
        \bottomrule
    \end{tabular}
\end{table}

\subsection{Training and network setup} \label{sec:tns}

In this subsection, we describe both the training and network setup of \proposal. First, we explain how our datasets are split. Then, the network setup is presented. Lastly, we discuss the parameters of \proposal.

In this study, we use six different datasets in our experiments. We need to split the images of these datasets into a train and test subsets, as the evaluation of \proposal requires it. We have created a 5-fold cross-validation evaluation environment, meanwhile for the rest of the datasets, their train and test subsets had already been predefined.

Another component of \proposal is the used network along all the experiments. In our case, we have chosen ResNet-50 as the pre-trained network. \begin{idea} We have selected ResNet-50 as the baseline feature extraction method following up the conclusions drawn in \citep{evoprunedeeptl}, in which several experiments with other feature extractors such as DenseNet and VGG were found to perform worse than ResNet-50. Although other larger feature extractors may provide better performance, the choice of ResNet-50 is also related to the number of features obtained from the network, which directly influences the evaluation time of a solution.  Based on these criteria, ResNet-50 is established as the pretrained backbone for our experiments, given its good balance between performance and complexity.\end{idea} This election has been taken to maintain the balance between the number of features, which leads to a higher computational space, and the performance obtained in the TL process. The combinatorial problem can be huge for typical values of feature extractors commonly used in problems where TL is in use. Using this network yields feature vectors $F_\phi(\mathbf{x})$ comprising $P=2,048$ components, leading to a total of $2^{2,048} \approx 3.23\cdot 10^{616}$ possible pruning patterns. Furthermore, the evaluation of pruned networks during the search requires repeatedly training over the instances in the test subset can be computationally expensive. Note that, although in our experiments a CNN model is used, the pruning can also be performed with other type or architectures, like Long Short-Term Memory (LSTM) \citep{lstmpruning}.

These extracted features are passed through the last layers, which are the layers that are going to be trained. The model with the larger accuracy on the training set is saved. The optimizer of the training environment is the standard SGD. The parameters of \proposal are shown in Table \ref{tbl:parameters}. The maximum number of training epochs is 600, but the training phase stops if no improvement is achieved in ten consecutive rounds. The last important parameter appears in the OoD phase. It is called \textit{Temp$_{ODIN}$} and it controls how the softmax values are computed using the logits from the Ind and OoD.  The parameters of this study (see Table 2) have been selected by following recommendations of the authors. The values of the parameters controlling the genetic search operators have been taken from \citep{evoprunedeeptl}. The characteristics of the neural network are also those utilized in this previous work. Moreover, the OoD detection mechanism is based on \citep{liang2017odin}. In that work different temperature values were tested, reporting the value of the parameters of the technique that rendered the best results in their experiments. For this reason, in this work we have chosen the same value (namely, temperature equal to 1000). 

\begin{table}[!h]
  \centering 
  \caption{Parameters of \proposal.}
  \label{tbl:parameters}
  \begin{tabular}{cc}
    \toprule
    \textbf{Parameter} & \textbf{Value}\\
    \midrule
    Maximum Evals & 200 \\
    \# Runs & 10 \\
    Population size &  30\\
    $p_{mut}$ & $\frac{1}{P}$\\
    Batch Size & 32 \\
    Temp$_{ODIN}$ & 1000 \\
    \bottomrule
  \end{tabular}
\end{table}

The last contribution of this section is the discussion of the parameters of \proposal. The maximum evaluations of \proposal is set to 200 and the size of the population of networks for each generation is 30. Table \ref{tbl:alltimeproposal} shows the evaluation time for each individual, so that the total time of execution is the time of the first column multiplied by the number of evaluations. Each OoD detection requires a minute, but in the datasets with the 5-fold cross-validation, this time reaches the five minutes. Moreover, we also indicate the inference time for test and the required time to calculate the AUROC metric in the OoD phase. Those times force us to keep a low number of runs and evaluations to meet a computationally affordable balance between the performance of our models and the high execution times required for our simulations. Moreover, although statistical tests are important to assess the significance of the differences in the results, but due to these limited number of runs, we can not apply them, as large number of runs is required to achieve statistically reliable insights.
\begin{table}[!h]
  \centering
  \caption{Average time in evaluations of \proposal.}
  \label{tbl:alltimeproposal}
  \begin{tabular}{crrrr}
    \toprule
    Dataset & Total & Evaluation & Training and Inference & OoD Detection \\
    \midrule
    CATARACT & 332 min & 1.66 min & 0.66 min & 1 min \\
    LEAVES & 1600 min & 8 min & 3 min & 5 min \\
    PAINTING & 1700 min & 8.5 min & 7.5min & 1 min \\
    PLANTS & 800 min & 4 min & 3 min & 1 min \\
    RPS & 900 min & 4.5 min & 3.5 min  & 1 min \\
    SRSMAS & 1500 min & 7.5 min & 2.5 min & 5 min \\
    \bottomrule
  \end{tabular}
\end{table}

The experiments have been carried out using Python 3.6 and a Keras/Tensorflow implementation deployed and running on a Tesla V100-SXM2 GPU.


\section{Results and discussion} \label{sec:results}

This section is devised to analyze the behavior of \proposal. To this end, we define three research questions (RQ) which are going to be answered in the following subsections with diverse experiments over the previous datasets. We will show and analyze several plots to illustrate the benefits of \proposal. The RQ can be stated as follows:
\begin{enumerate}[leftmargin=1.2cm, start=1,label={(\bfseries RQ\arabic*)}]
    \item How are the approximated Pareto fronts produced by the proposal in each of the considered datasets? \newline

    The Pareto front can be defined as the set of non-dominated solutions, being chosen as optimal, if no objective can be improved without sacrificing at least one other objective. The problem at hand is approximated using a multi-objective approach. For that reason, we want to check that not only we have promising solutions in the extreme values of the Pareto front, but also to have a wide population of diverse solutions in the whole Pareto. As a consequence of that, to answer this RQ, we will analyze how is the Pareto front for each dataset and if there exists any direct connection between the objectives of the study: accuracy, complexity of the network, and robustness.  In addition, a comparison to other pruning methods from the literature will be performed to check whether our proposal performs competitively against such methods.

    \item Is there any remarkable pruning pattern that appears in all the solutions of the Pareto front? \newline

    We compare the pruning patterns of all the models of the Pareto fronts of \proposal to show if there are some important patterns which are key to identify the most important zones of the input images. We employ a well-known technique called Grad-CAM \citep{gradcam}, which uses the gradient of the classification score with respect to the convolutional features of the network to check which parts of the image are most important for the classification task. Grad-CAM lies in the group of Explainable Artificial Intelligence (XAI) techniques, as it produces details to make easy to understand which neurons are the relevant ones in all the experiments \citep{xaibarredo}. These neurons lead to specific pixels or group of them of the original images that are passed through the network.

    \item Do our models achieve an overall improvement in performance when merged through ensemble modeling? \newline

      \proposal trains a great variety of models which leads to a wide diversity of models in the Pareto front for each dataset. The aim of this RQ is to check whether the diversity of pruning patterns in the Pareto front can be used to improve our DL models through ensemble strategies. Our aim is to check if an ensemble of differently pruned models can yield more accurate predictions, leading to a better overall performance than their compounding models in isolation. Beyond improving the accuracy through ensembles, we will also explore whether ensemble modeling allows obtaining more robust models, so that the number of OoD samples that the network wrongly predicts as InD is lower.

\end{enumerate}

This section is divided in Section \ref{sec:rq1}, where we analyze the different Pareto front for each considered dataset in order to answer RQ1. Next, in Section \ref{sec:rq2}, we will examine the different pruning patterns of our models. Precisely, we will look for the neurons that appears in most of them, and we will highlight the essential zones of the input images, as this is the key part to answer RQ2. Lastly, we will discuss in Section \ref{sec:rq3} the benefits of the diversity of our models when ensemble modeling is performed, to show if an improvement in terms of accuracy and AUROC is achieved, which the principles lines of the RQ3.

\subsection{Answering RQ1: Analyzing the Pareto fronts of \proposal} \label{sec:rq1}

The objective of this section is to answer RQ1 by performing a complete analysis of the Pareto fronts of \proposal and then, performing a comparison against competitive pruning methods of the literature.  This analysis is to be performed focusing on two important aspects: i) how are the Pareto fronts for each dataset? and ii) how are the projections in each objective for each dataset? \proposal is run for 10 times, each yielding an estimation of the Pareto front between the three objectives. Such Pareto front estimations contain solutions that dominate -- in the Pareto sense -- the rest of evaluated solutions during the evolutionary search. The elitist nature of the algorithm ensures that non-dominated solutions are retained in the population. Moreover, after the 10 executions of the algorithm, all Pareto front estimations are merged together. Non-dominated solutions in this merger compose a new Pareto front estimation (i.e., a \emph{super} Pareto front) containing the best solutions found across the 10 runs of the algorithm. For simplicity, these solutions will be hereafter denoted as the Pareto front discovered by \proposal. Moreover, for each Pareto front, it has been included the results for the non-pruned network for each dataset, which are composed of solutions with all the active neurons and the accuracy and AUROC showed in Table \ref{tbl:datasets}.

\begin{figure}[!hbtp]
    \includegraphics[width=\linewidth]{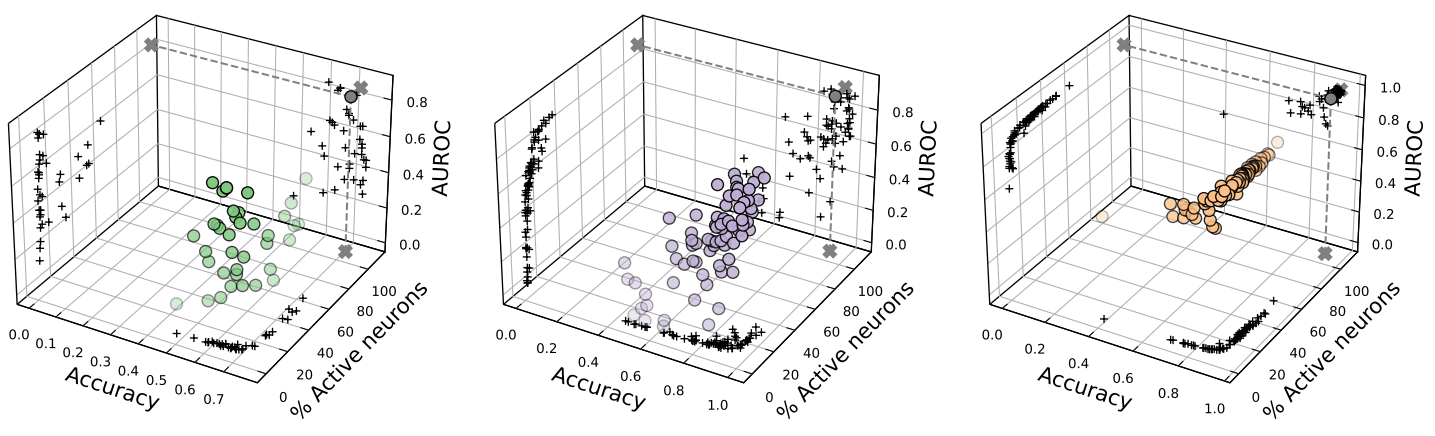}
    \caption{Pareto fronts of \proposal. Left: CATARACT dataset. Middle: RPS dataset. Right: PAINTING dataset.}
    \label{fig:paretos1}
\end{figure}

With these graphics we analyze the quality of each Pareto and, particularly, by assessing the full spectrum of solutions that can be achieved in each of the Pareto. Moreover, we are going to study whether there is a direct relationship between any of the objectives we have formulated in the previous sections. In order to develop these plots, we have collected all the solutions of the super Pareto front (called Pareto front from now), selecting 10\% of the best solutions for each objective in order to make their projections.

\begin{figure}[!hbtp]
    \includegraphics[width=\linewidth]{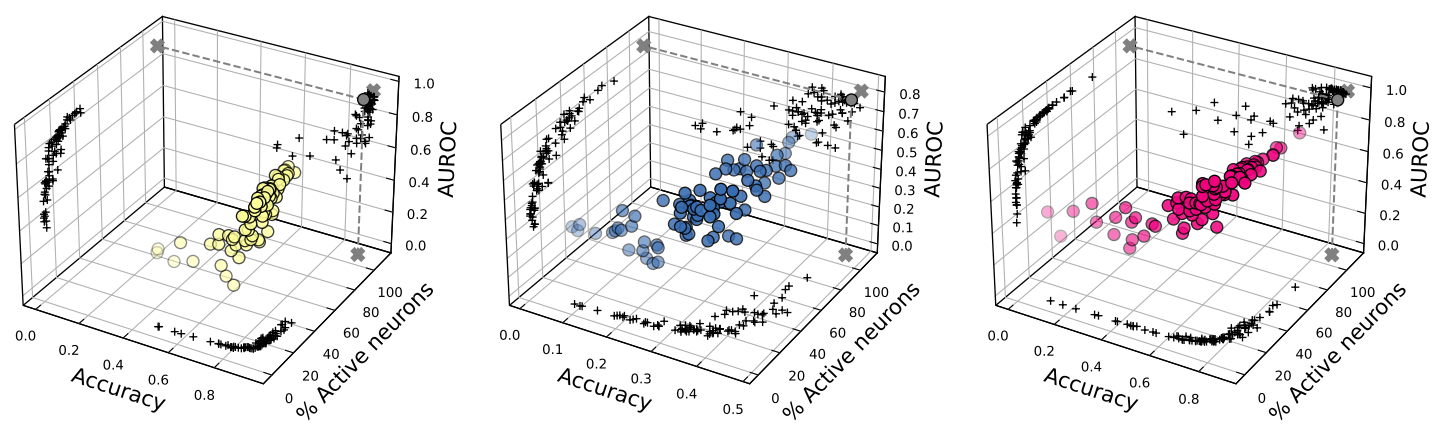}
    \caption{Pareto fronts of \proposal. Left: LEAVES dataset. Middle: PLANTS dataset. Right: SRSMAS dataset.}
    \label{fig:paretos2}
\end{figure}

These Pareto front are presented in Figures \ref{fig:paretos1} and \ref{fig:paretos2}. We can observe the diversity of pruning patterns produced by \proposal. Moreover, another insight from these Pareto front comes up when we inspect extreme values of each objective, as they systematically achieve good results in each dataset. Most of the solutions obtain high values of accuracy and robustness meanwhile their remaining active neurons are kept low.

First, we focus on the central part of the 3D projections, in which we visualize the three objectives. Our goal is to detect if there exist some kind of relationship between them. We clearly see that the projection in all the Pareto front takes values to the upper corner in which the three objectives present low values of percentage of active neurons, but high accuracy and AUROC. Moreover, this distribution of the points in both group of figures indicate that there is a tendency of the solutions to that plane in which the number of active neurons is low.

Analyzing the two-dimensional planes, there is not a clear relation between the performance and robustness. Nonetheless, the common point of this projection, namely, the complexity of the network, sheds light to the fact that it can be related with the performance and robustness separately. In both cases, a minimum number of active neurons is needed in order to start achieving good results in each objective. For both objectives, there is a certain range of optimal number of active neurons in which each of them obtains their best values.

The last experiment of this section addresses the performance comparison between \proposal and other competitive pruning methods from the related literature. In this work, we compare \proposal to the following two methods \begin{idea} considered to be competitive counterparts for benchmarks between pruning proposals \citep{Hoefler2021}\end{idea}:
    \begin{itemize}[leftmargin=*]
        \item \texttt{weight} \citep{han2015learning}: The parameters with lower values are pruned at once. This method operates over the whole parameter set in the layer to be optimized.
        \item \texttt{neuron} \citep{srinivas2015data}: The neurons with lower mean input connection values are pruned. 
    \end{itemize}

    

    \begin{idea}
        Both pruning methods require a parameter that controls their execution, which is the target percentage of remaining neurons. This percentage represents the active weights remaining in the network that each of these methods reaches at the end of its execution. For a fair comparison, we force these methods to target the same percentage of remaining weights as in the solutions of the Pareto front estimated by \proposal. We note that the Pareto front estimation contains the non-dominated solutions found in the 10 runs of the algorithm. In this case, we have ordered the solutions in terms of complexity (second objective), which yields a distribution of ordered solutions, from least to most active weights. Based on this sorted list of solutions, we have chosen those with the median and the lowest values of the percentage of the remaining active weights (complexity of the network) to assess how \proposal behaves in these representative cases. Once we annotate these percentages, the pruning methods prune the fully-connected network until reaching such annotated values, giving rise to the accuracy values shown in the columns of this table.  
    \end{idea}

    \begin{idea}
        Table \ref{tbl:pruningcomparison} shows the results of the comparison between \proposal and the pruning methods. The second column indicates the target percentage of remaining weights corresponding to each dataset. The third, fourth, and fifth columns report the accuracy of weight pruning, neuron pruning, and \proposal, respectively. In addition, each dataset spans two rows in the table: the first row shows the median percentage of active weights and the accuracy of each of the proposals for that case, whereas the second row represents the case with the lowest percentage of active weights and their respective levels of accuracy for each approach.
    \end{idea}
    
    \setcounter{table}{3}
    \begin{table}[!h]
        \centering        
        \caption{Comparison of \proposal against competitive pruning methods of the literature in terms of accuracy given a fixed value of {\color{blue}target percentage of pruning weights}.}
        \label{tbl:pruningcomparison}
        \begin{tabular}{lcccc}
        \toprule
        \multirow{2}{*}{Dataset} & {\color{blue} Percentage of} & \multirow{2}{*}{\texttt{weight}} & \multirow{2}{*}{\texttt{neuron}} & \multirow{2}{*}{\proposal} \\ 
         & {\color{blue}remaining weights} & & &  \\ \midrule
        \multirow{2}{*}{CATARACT} & 3.3\%     & 0.380 & 0.248  & \textbf{0.694}  \\ 
                                  & 0.09\%     & 0.182  & 0.165  & \textbf{0.504}  \\ \midrule
        \multirow{2}{*}{LEAVES}   & 10.80\%     & 0.637  & 0.700  & \textbf{0.906}  \\ 
                                  & 2.90\%     & 0.462  & 0.450  & \textbf{0.515}  \\ \midrule
        \multirow{2}{*}{PAINTING} & 15.5\%     & 0.747  & 0.524  & \textbf{0.935}  \\ 
                                  & 0.09\%     & 0.333  & 0.107  & \textbf{0.429}  \\ \midrule
        \multirow{2}{*}{PLANTS}   & 11.1\%     & 0.212  & 0.072  & \textbf{0.373}  \\ 
                                  & 0.25\%     & 0.043  & 0.034  & \textbf{0.106}  \\ \midrule
        \multirow{2}{*}{RPS}      & 5.0\%     & \textbf{0.943}  & 0.900  & 0.894  \\ 
                                  & 0.24\%     & \textbf{0.943}  & 0.333  & 0.484  \\ \midrule
        \multirow{2}{*}{SRSMAS}   & 8.79\%     & 0.454  & 0.408  & \textbf{0.782}  \\ 
                                  & 0.10\%     & \textbf{0.161}  & 0.079  & 0.145  \\ \bottomrule
        \end{tabular}
    \end{table}
    
    

    \begin{idea}
        Results in the above table evince that \proposal outperforms these pruning methods in most of the datasets. There are four datasets in which, without any doubt, \proposal achieves a better performance than pruning methods. For the SRSMAS dataset, \textit{weight} is slightly better than \proposal in the case of the lowest percentage of active weights. This difference might be enough to state that SRSMAS performs better than \proposal. Nonetheless, the median case shows that, when a minimal number of neurons/weights are active, our proposal outperforms \textit{weight} in this dataset. 
        
        A special case is noted in the results for the RPS dataset, which is the easiest one in terms of modeling difficulty. Results expose this fact because, when the approach is to eliminate a whole group of connections represented by the neurons, \proposal achieves a better performance in both cases. In fact, the greater the number of neurons/connections to be active is, the better both models will perform. However, if the strategy is to eliminate single connections as implemented by the \textit{weight} strategy, it does not imply removing the whole set of connections of the neuron. In this case, this method may perform better than \proposal. The fact that RPS is the simplest dataset is reflected in the fact that the same accuracy value can be achieved by several desired pruning configurations. Based on this observation, it can be concluded that removing connections is a valid pruning method especially when complemented with other techniques such as evolutionary algorithms. In this case, there is potential for improvement in extreme, intermediate or general cases, as shown in the Pareto front estimations reported in these results.        

        Results attained by \proposal at the median percentage of pruning neurons are remarkable, since it corresponds to the center of the distribution of complexity values in the Pareto front estimated by the technique. In detail, all cases report a minimum of approximately 70\% of pruned weights in the worst case. In the best case, almost the entire network is pruned, which corresponds to the lowest complexity value in the estimated front. The higher the percentage of pruned neurons is, the more difficult is to achieve a model with good accuracy levels, since a minimal amount of neurons/weights is needed to achieve them. This is exposed in most considered cases, in which the median value achieves a better performance. Taking a closer look at the case with lowest percentage of remaining weights, which can be deemed a more complex case, the performance of the models degrades, which is one of the lessons learned from the inspection of the Pareto fronts made in this section. However, \proposal is able to outperform the rest of pruning methods with models that do not surpass 3\% of active neurons, except in the case of SRSMAS, whose performance is practically the same.
\end{idea}

The Pareto fronts shown in this section have allowed us to obtain valuable information on the different executions of \proposal. The configuration of \proposal has allowed us to obtain a fairly diverse set of solutions, with competitive solutions at the extreme values of the different objectives of the study. A second conclusion drawn from this analysis is the existence of relationship or direct Pareto both the complexity of the network and its performance and the complexity and robustness, but it does not appear to exist between the performance and the robustness. Finally, a third conclusion has been drawn from the comparison against other pruning methods: in general, \proposal is able to outperform such methods for both intermediate and extreme pruning values, whereas a minimum percentage of neurons is required to produce high-quality pruned models. 

\subsection{Answering RQ2: Remarkable pruning patterns in the Pareto fronts of \proposal} \label{sec:rq2}

This RQ aims to analyze if there are certain pruning patterns, along the different trained networks, that allows detecting important regions in the input images to the pruned networks.

In order to answer this RQ, we must discriminate relevant neurons that appear in most of the pruning patterns in the Pareto fronts produced by \proposal. In doing so, we resort to a XAI technique called GradCAM \citep{gradcam}, which permits to localize the regions within the image that are responsible for a class prediction. Thanks to GradCAM, we can go backwards from the neurons of the solutions and highlight these key pixel regions. For each dataset, we depict several query images, and remark the 10 most relevant neurons as per GradCAM and their distribution among the three objectives. In the following figures, the central sections are relevance heatmaps obtained by GradCAM, remarking the most influential zones of the input images as warmer colors. In addition to the heatmap, we also present two more plots. The first one, in the left top, show as a bar diagram the index of the 10 most relevant neurons which appear as active in most of the solutions of the Pareto front, with their relative frequency. The second one, at the right, shows the distribution of the objective's values for these representative neurons. In this chart, a boxplot is shown for each objective and neuron: from left to right, accuracy, percentage of active neurons and AUROC, respectively. The border of the heatmaps and the color of the bars in the figures are related, so that the reader can match each heatmap to the corresponding neuron and frequency of appearance in the Pareto.

\begin{figure}[!hbtp]
    \centering
    \subfloat[Bars of CATARACT]{\includegraphics[width= 0.4\textwidth]{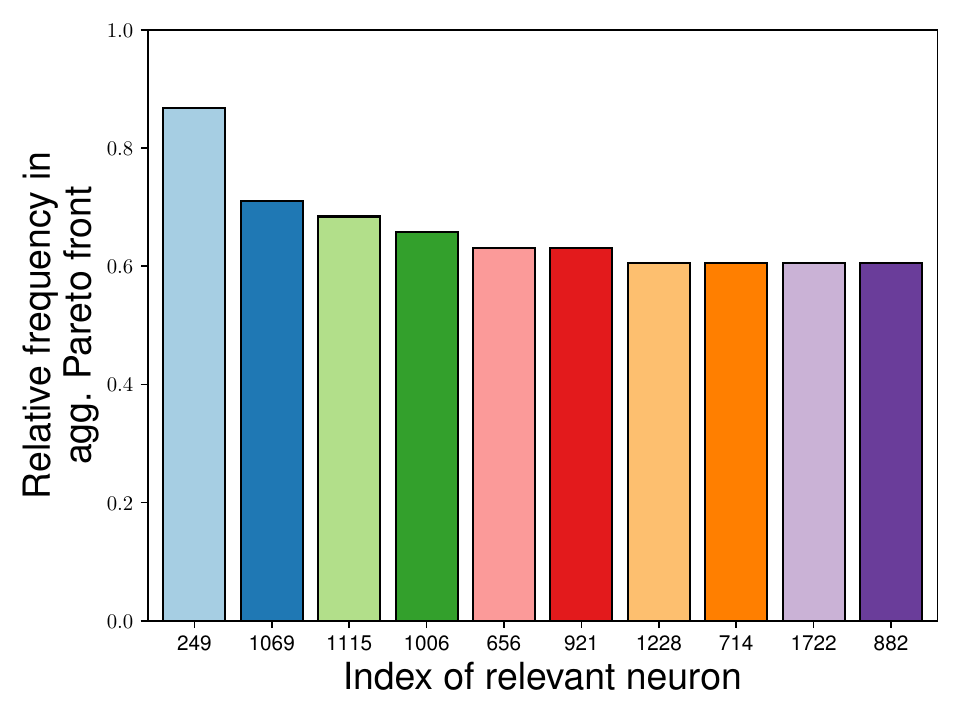}}
    \subfloat[Boxplots of CATARACT]{\includegraphics[width= 0.4\textwidth]{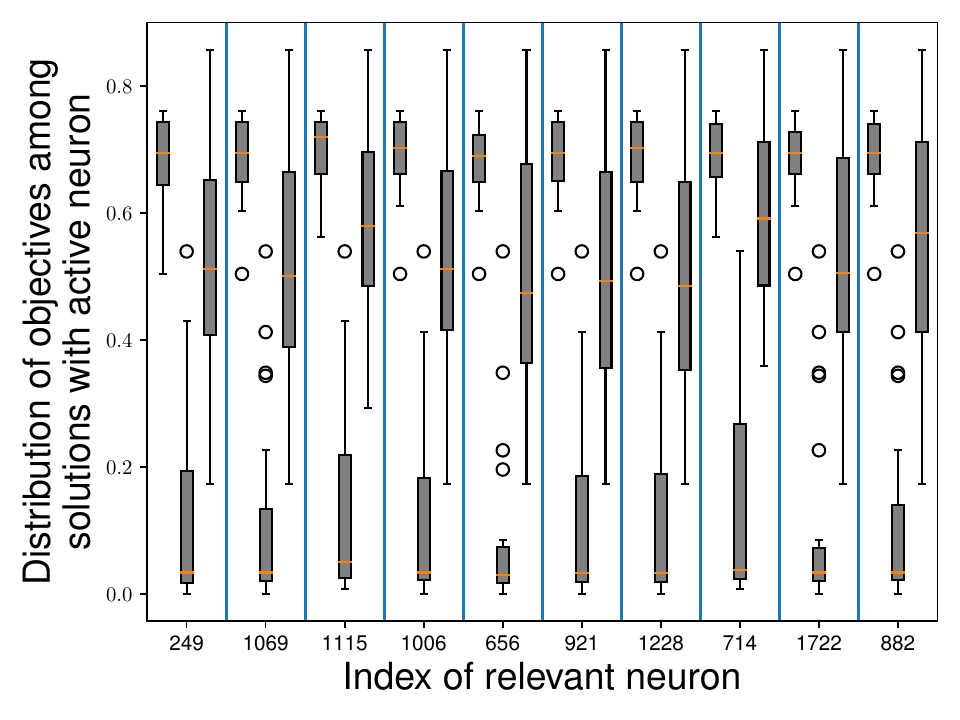}}\\
    \subfloat[Heatmaps of CATARACT]{\includegraphics[width= 0.9\textwidth]{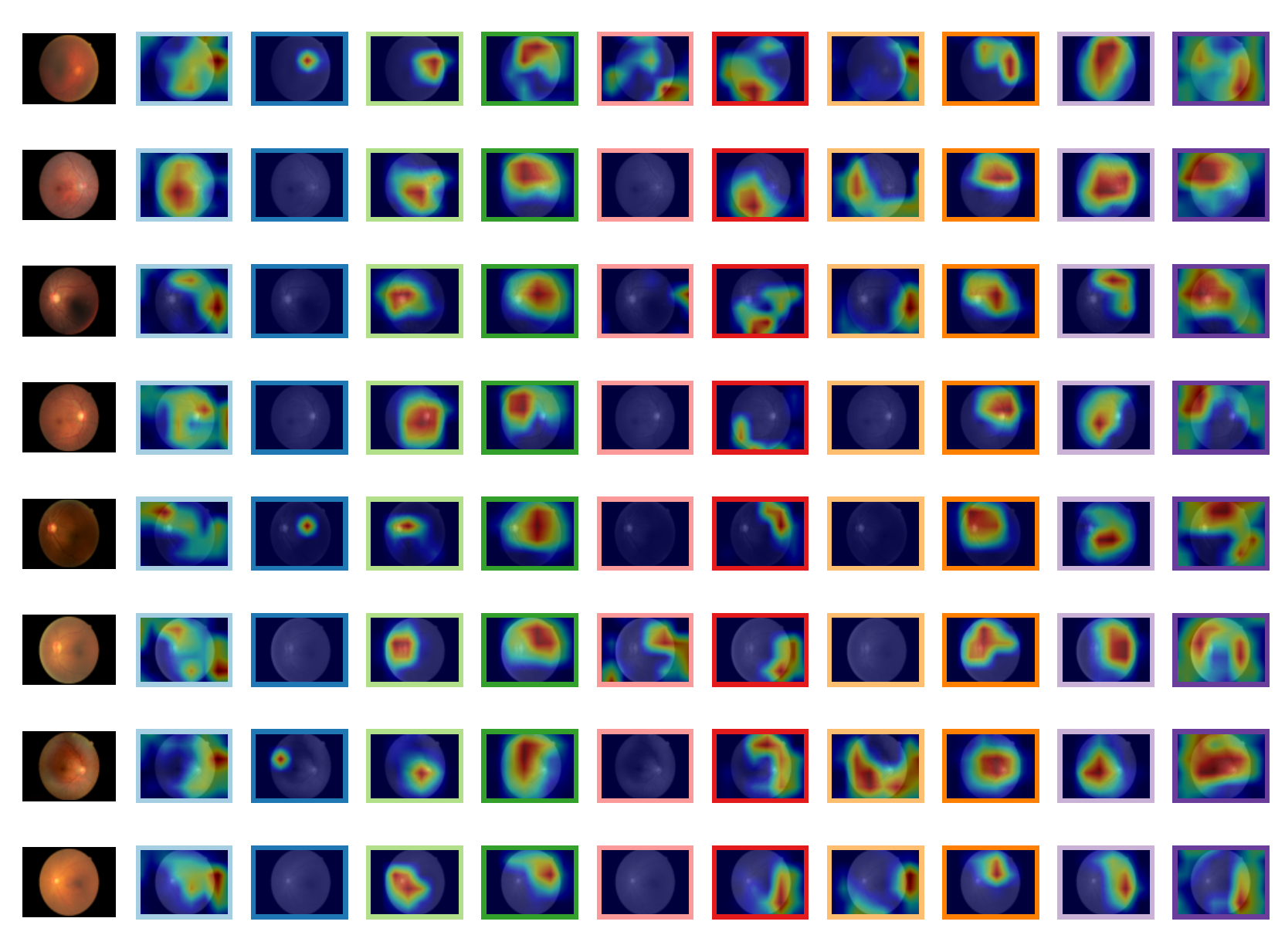}}
    \caption{Bars, boxplots and heatmaps of CATARACT.}
    \label{fig:xaicataract}
\end{figure}

Figure \ref{fig:xaicataract} shows the previous information for the CATARACT dataset. In the first one, the barplot, we can see that the least important neuron achieves a 60\% of frequency in the Pareto front, i.e., it appears in the 60\% of solutions meanwhile the best one has a frequency rate of more than the 80\%. The second figure, the boxplot, shows the distribution of the objectives for solutions which have these relevant neurons. These results show that low complexity is presented in these neurons and high accuracy and AUROC. Lastly, we see the heatmaps for this dataset. For the shown images, we can see how these pruning patterns that \proposal achieves during its evolutionary process. These patterns let us recognize how the network dictate the class for each image thanks to these ten most important neurons.

The next figure, Figure \ref{fig:xairps} shows the results for the RPS dataset. The bar graph shows that these neurons achieve an appearance in more than the 80\% of solutions of the Pareto and the boxplot confirm that the solutions in which these neurons are presented achieve, in most cases, less than 10\% of active neurons, accuracies near 90\% and AUROC around an 80\%. The examples images shown in the heatmaps present the effect of these important neurons. As we have previously noted, the keys to recognize the images are position of the fingers and even the separation among them, as warmer color are presented for them.

\begin{figure}[!hbtp]
    \centering
    \subfloat[Bars of RPS]{\includegraphics[width= 0.4\textwidth]{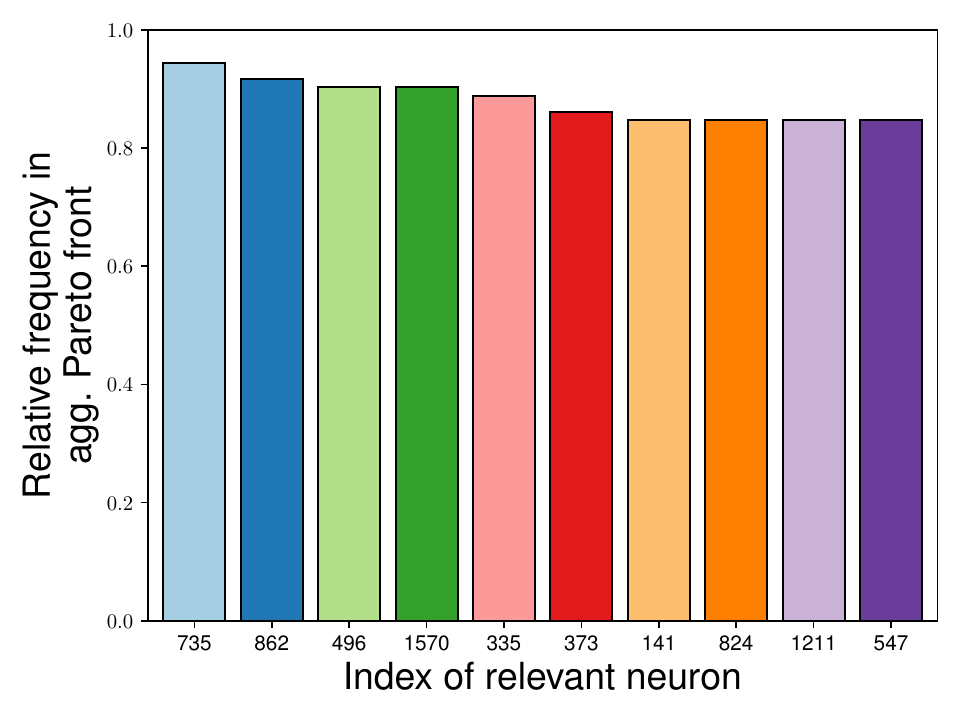}}
    \subfloat[Boxplots of RPS]{\includegraphics[width= 0.4\textwidth]{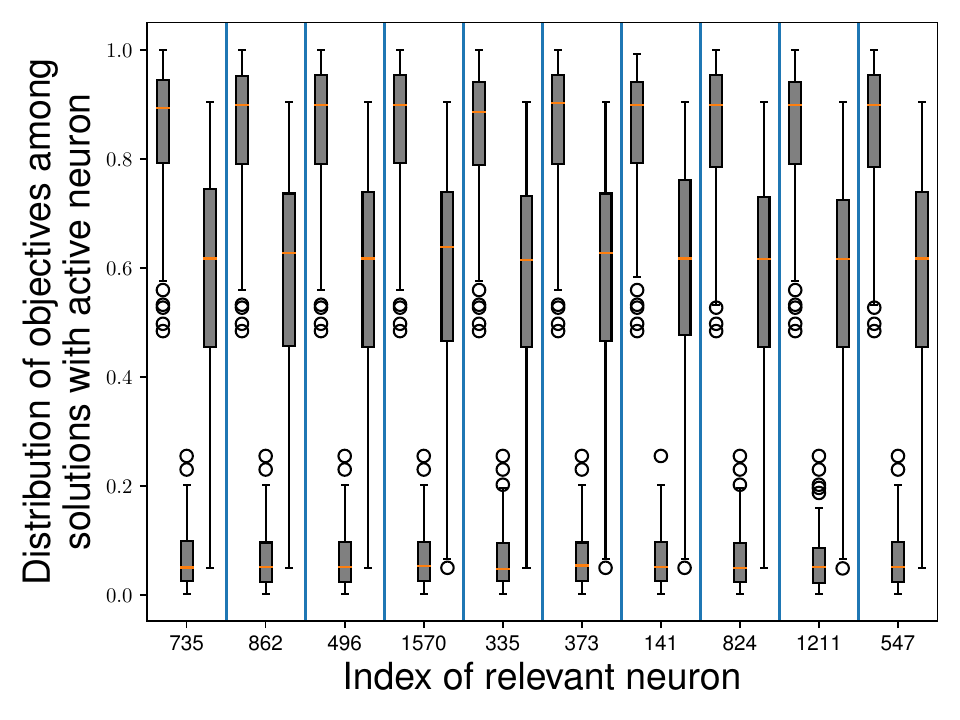}}\\
    \subfloat[Heatmaps of RPS]{\includegraphics[width= 0.9\textwidth]{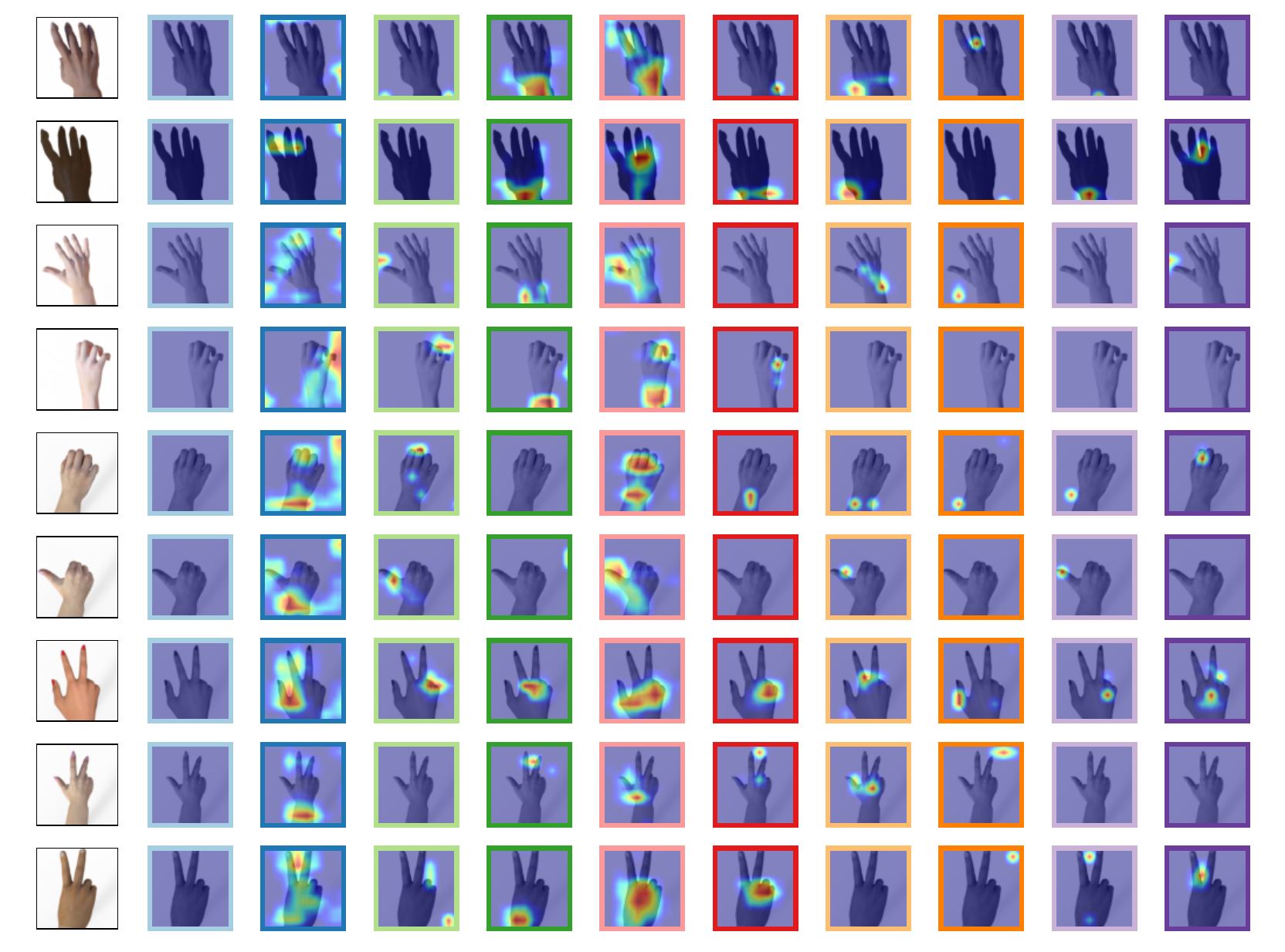}}
    \caption{Bars, boxplots and heatmaps of RPS.}
    \label{fig:xairps}
\end{figure}

\begin{figure}[!hbtp]
    \centering
    \subfloat[Bars of PAINTING]{\includegraphics[width= 0.4\textwidth]{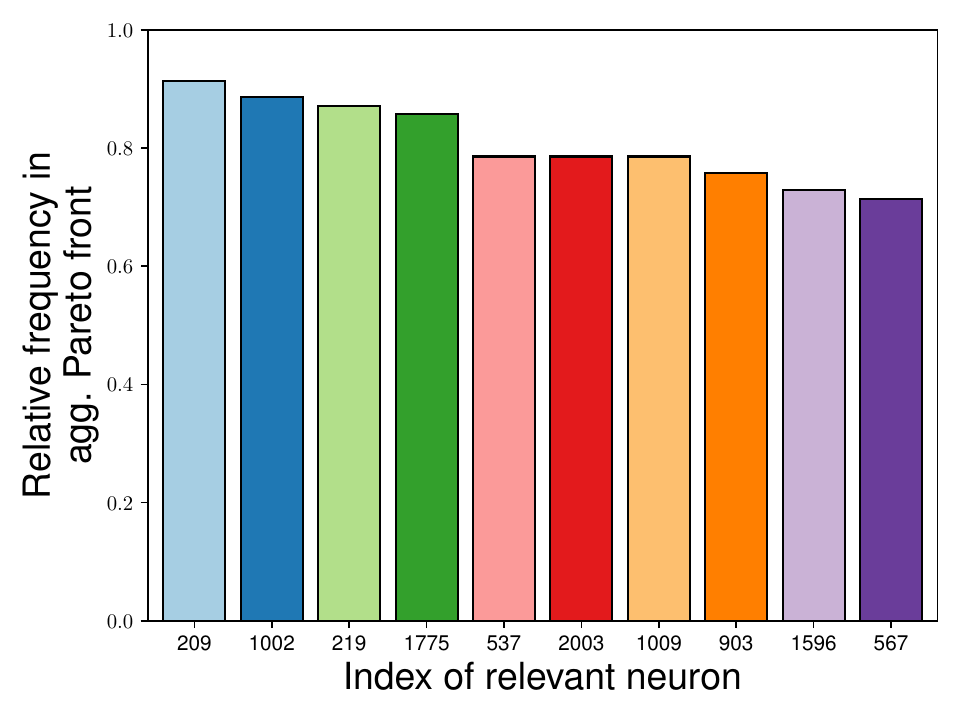}}
    \subfloat[Boxplots of PAINTING]{\includegraphics[width= 0.4\textwidth]{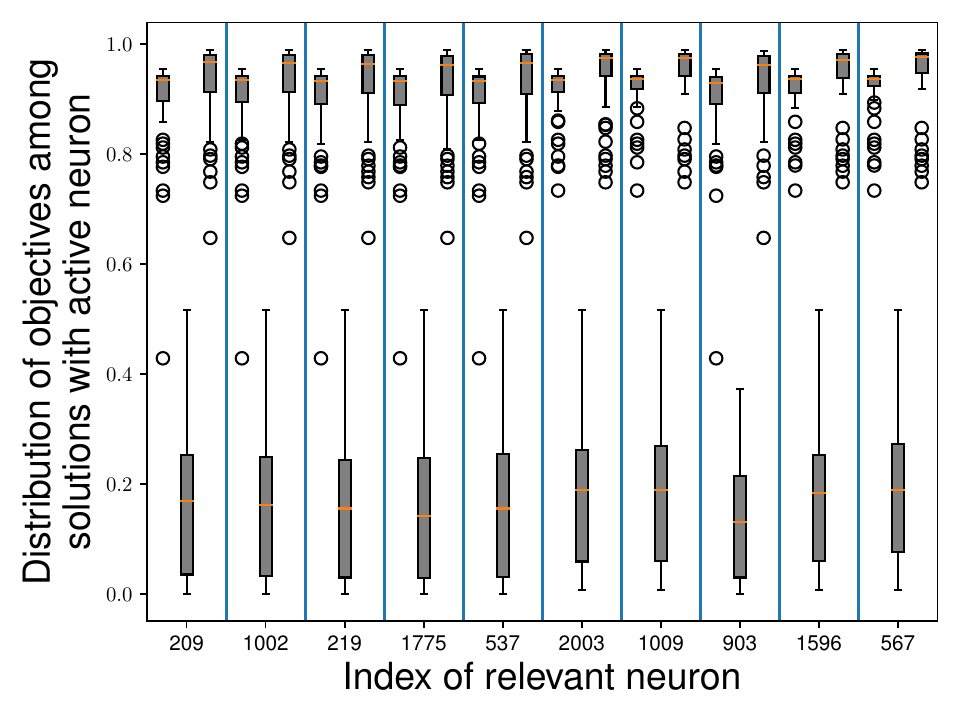}}\\
    \subfloat[Heatmaps of PAINTING]{\includegraphics[width= 0.9\textwidth]{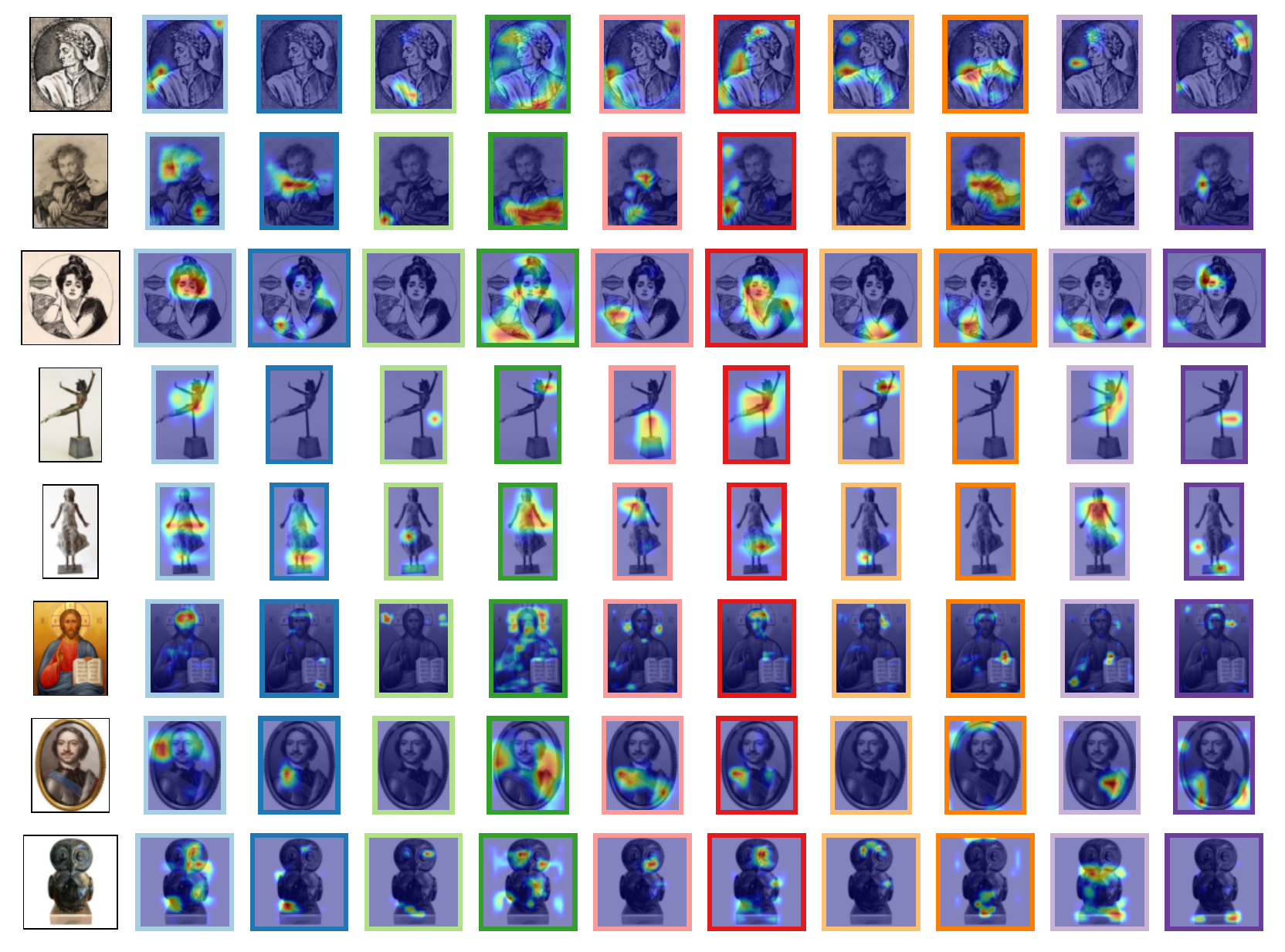}}
    \caption{Bars, boxplots and heatmaps of PAINTING.}
    \label{fig:xaipainting}
\end{figure}

The next dataset is PAINTING. Figure \ref{fig:xaipainting} shows the set of graphics for this dataset. The relevant neurons for PAINTING achieve a minimum percentage of appearance of 70\% among all the solutions in the Pareto front. There is a significant difference between the first relevant neuron and the rest in terms of appearance.  These solutions present almost a 20\% of active neurons, but also high performance both in accuracy and AUROC, between 90 and 100\%. This indicates the great level of uniformity in the robustness for this dataset. These neurons help us to analyze the images of this dataset. The third image presents a woman and, taking a deep look into the heatmaps, we see that the network recognizes the face, and then the outer parts, like the arms and the hair. Another interesting image is the fifth one. Our network is able to recognize the chest and also the arms and the rest of the body extremities.

\begin{figure}[!hbtp]
    \centering
    \subfloat[Bars of PLANTS]{\includegraphics[width= 0.4\textwidth]{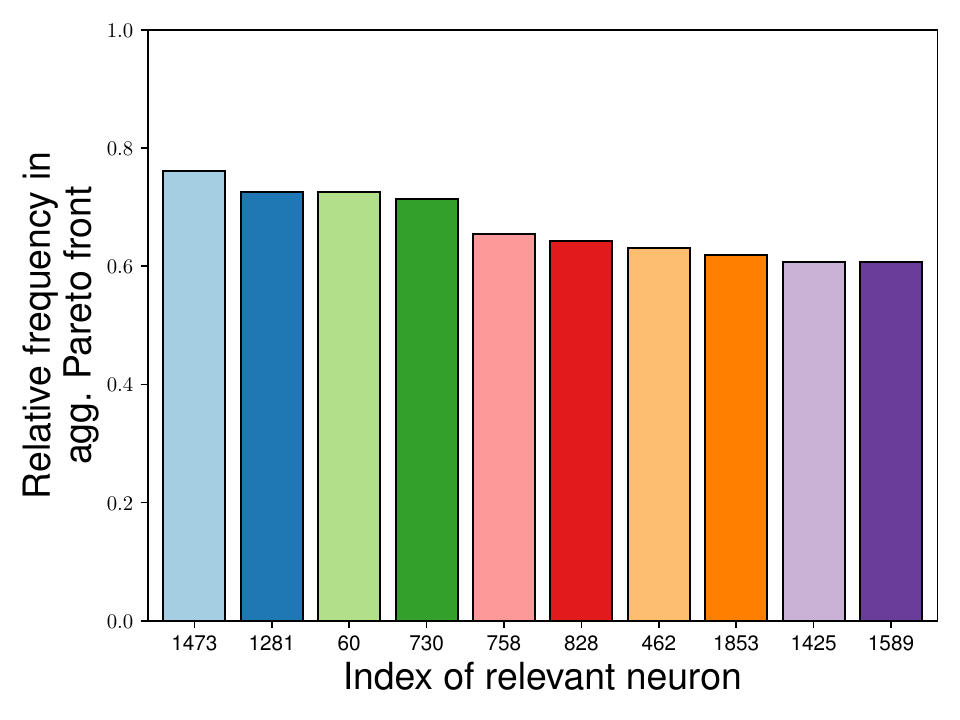}}
    \subfloat[Boxplots of PLANTS]{\includegraphics[width= 0.4\textwidth]{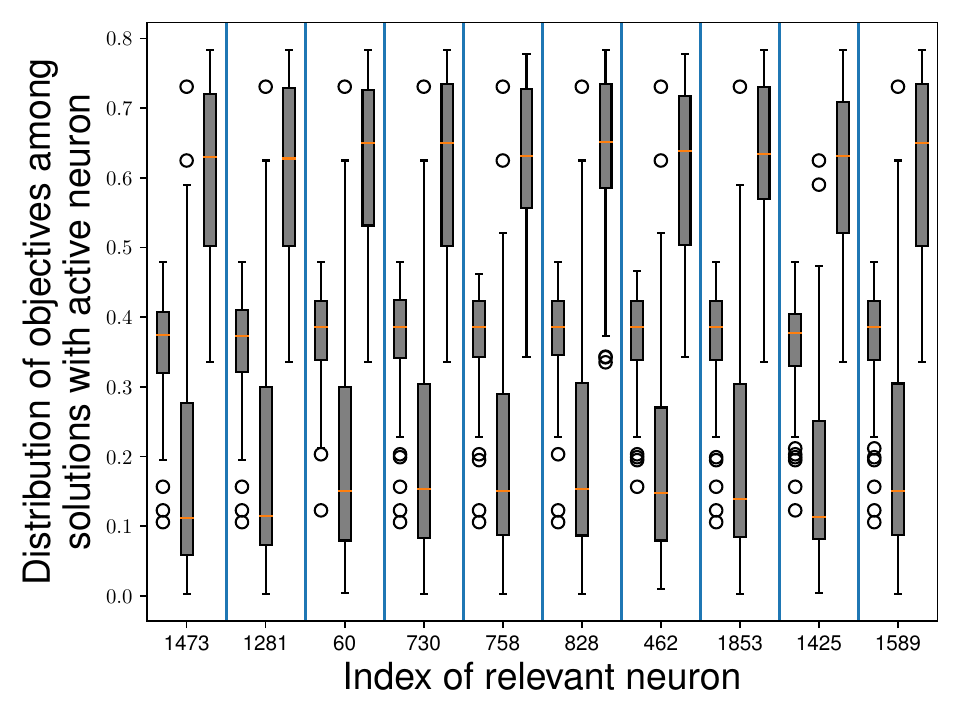}}\\
    \subfloat[Heatmaps of PLANTS]{\includegraphics[width= 0.9\textwidth]{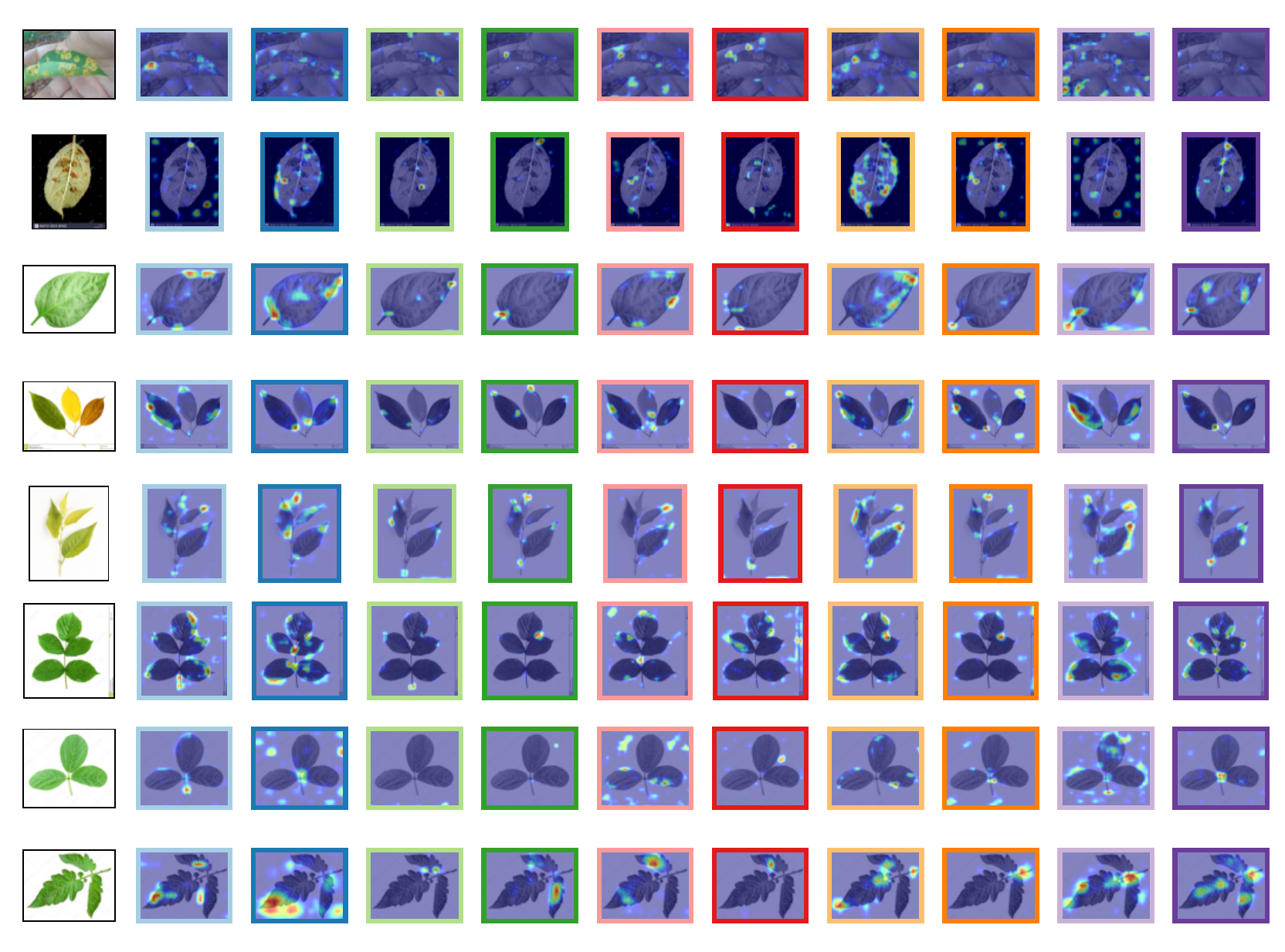}}
    \caption{Bars, boxplots and heatmaps of PLANTS.}
    \label{fig:xaiplants}
\end{figure}

We continue our analysis of the obtained pruning patterns of \proposal with the PLANTS dataset, shown in Figure \ref{fig:xaiplants}. The most important neurons have an appearance rate between 60 and 80\% in all the solutions of the Pareto front. Their distribution of objective report us a very low complexity of the network, near the 10\% in average, with a good result for this dataset both in accuracy and AUROC. This dataset contains images of leaves and plants of fruit and vegetables and, for that reason, our network focus in the recognition of the shape of these leaves, as it is shown in the three bottom images of the figure.

\begin{figure}[!hbtp]
    \centering
    \subfloat[Bars of LEAVES]{\includegraphics[width= 0.4\textwidth]{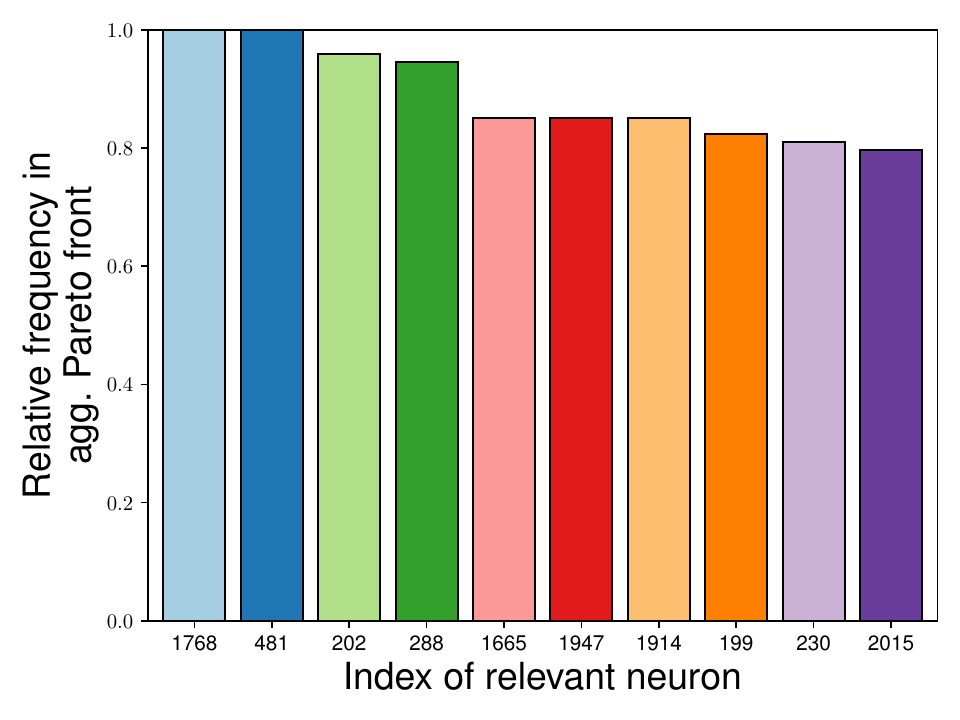}}
    \subfloat[Boxplots of LEAVES]{\includegraphics[width= 0.4\textwidth]{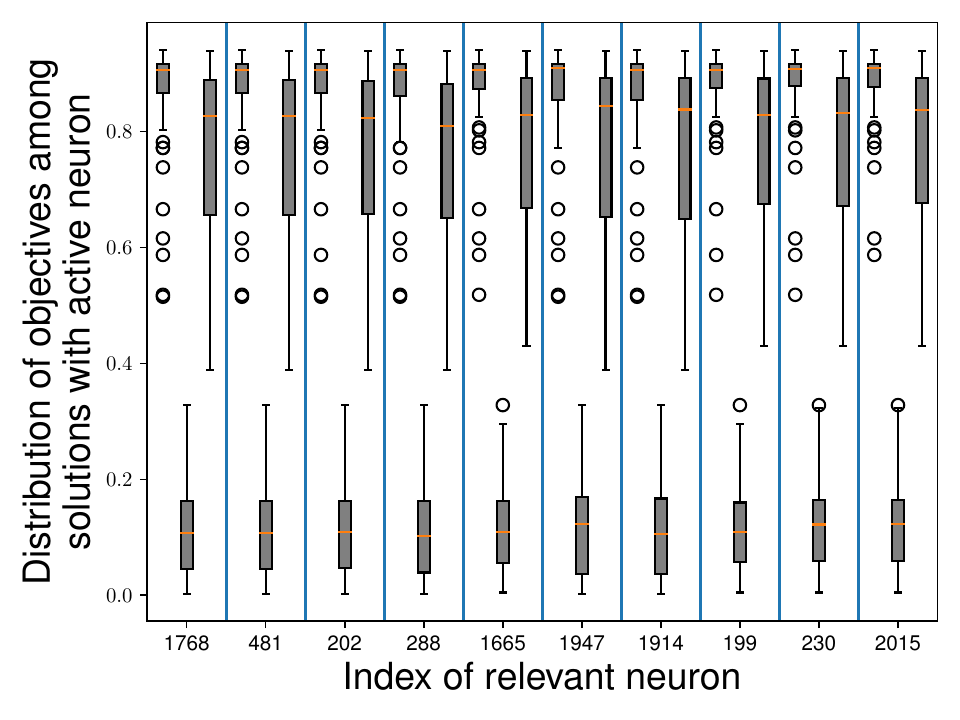}}\\
    \subfloat[Heatmaps of LEAVES]{\includegraphics[width= 0.9\textwidth]{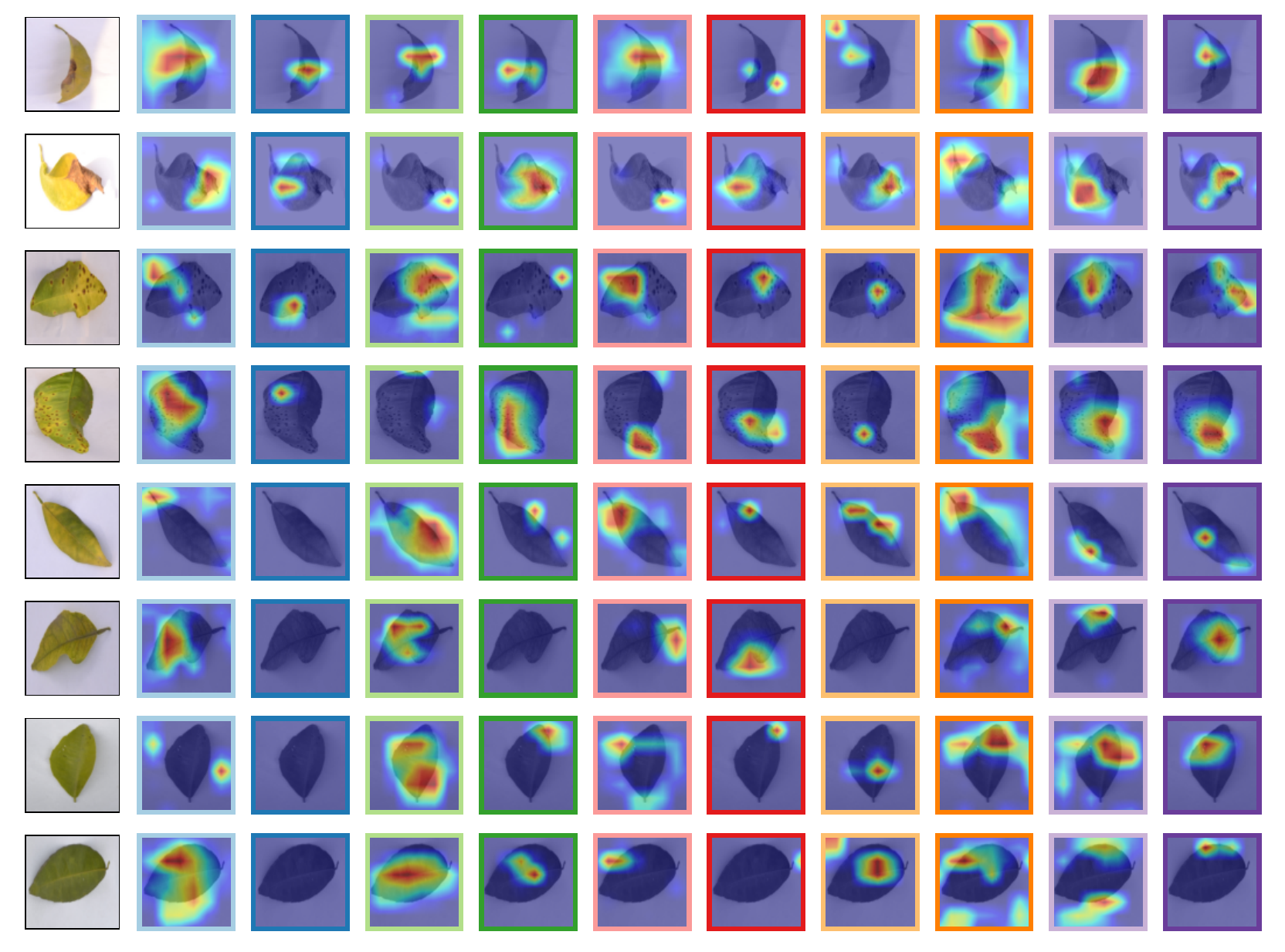}}
    \caption{Bars, boxplots and heatmaps of LEAVES.}
    \label{fig:xaileaves}
\end{figure}

We continue with this analysis with the LEAVES dataset. In this case, Figure \ref{fig:xaileaves} shows the three graphics for this dataset. The first one, which is related with the relevance of the neurons, exhibit two neurons which appear in all the solutions of the Pareto front and another two which present almost a 100\% of appearance. For those neurons, the boxplot chart report us similar distributions because, in all the cases, the remaining active neurons are kept low and the accuracy and AUROC are high. Lastly, the images from this dataset show both diseased and healthy leaves. The achieved pruning patterns of \proposal are able to distinguish the healthy from the diseased leaves (last image versus the third one starting from the top), and then the type of the disease.

\begin{figure}[!hbtp]
    \centering
    \subfloat[Bars of SRSMAS]{\includegraphics[width= 0.4\textwidth]{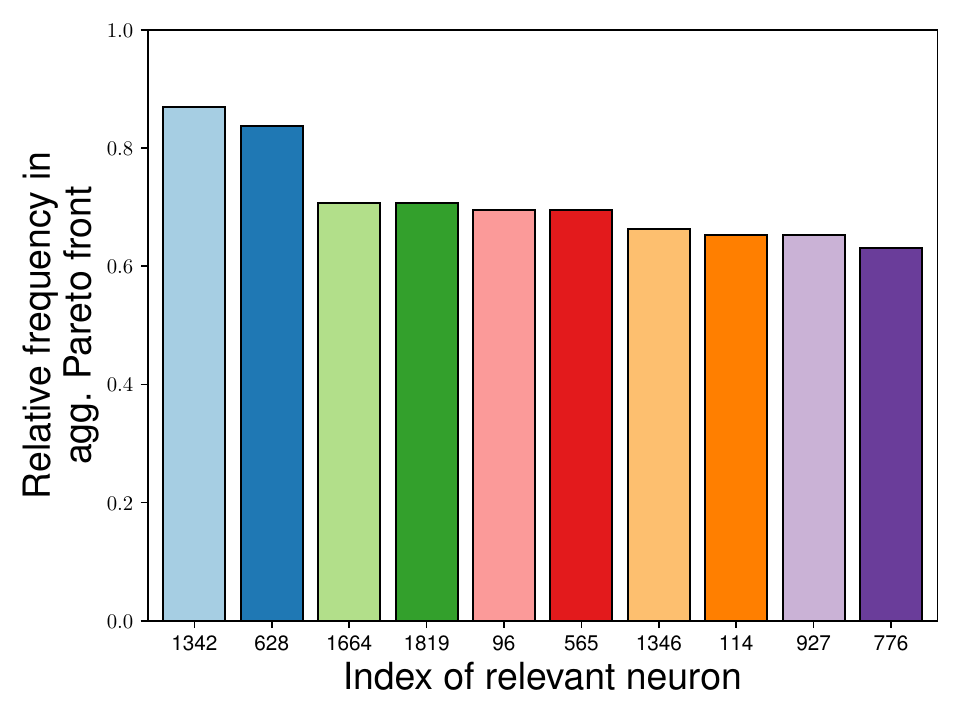}}
    \subfloat[Boxplots of SRSMAS]{\includegraphics[width= 0.4\textwidth]{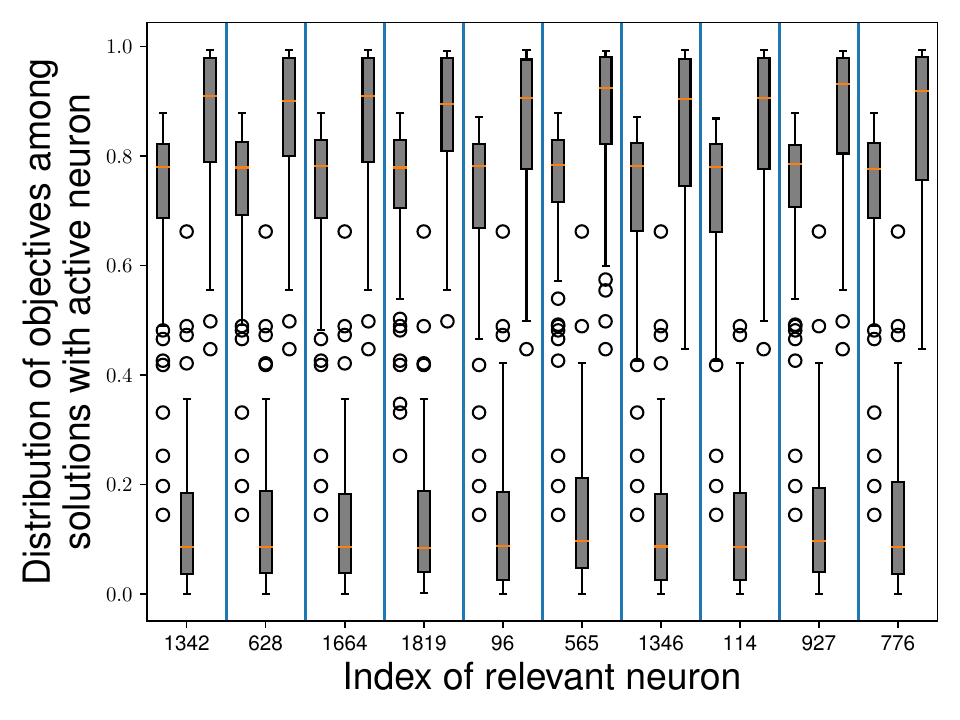}}\\
    \subfloat[Heatmaps of SRSMAS]{\includegraphics[width= 0.9\textwidth]{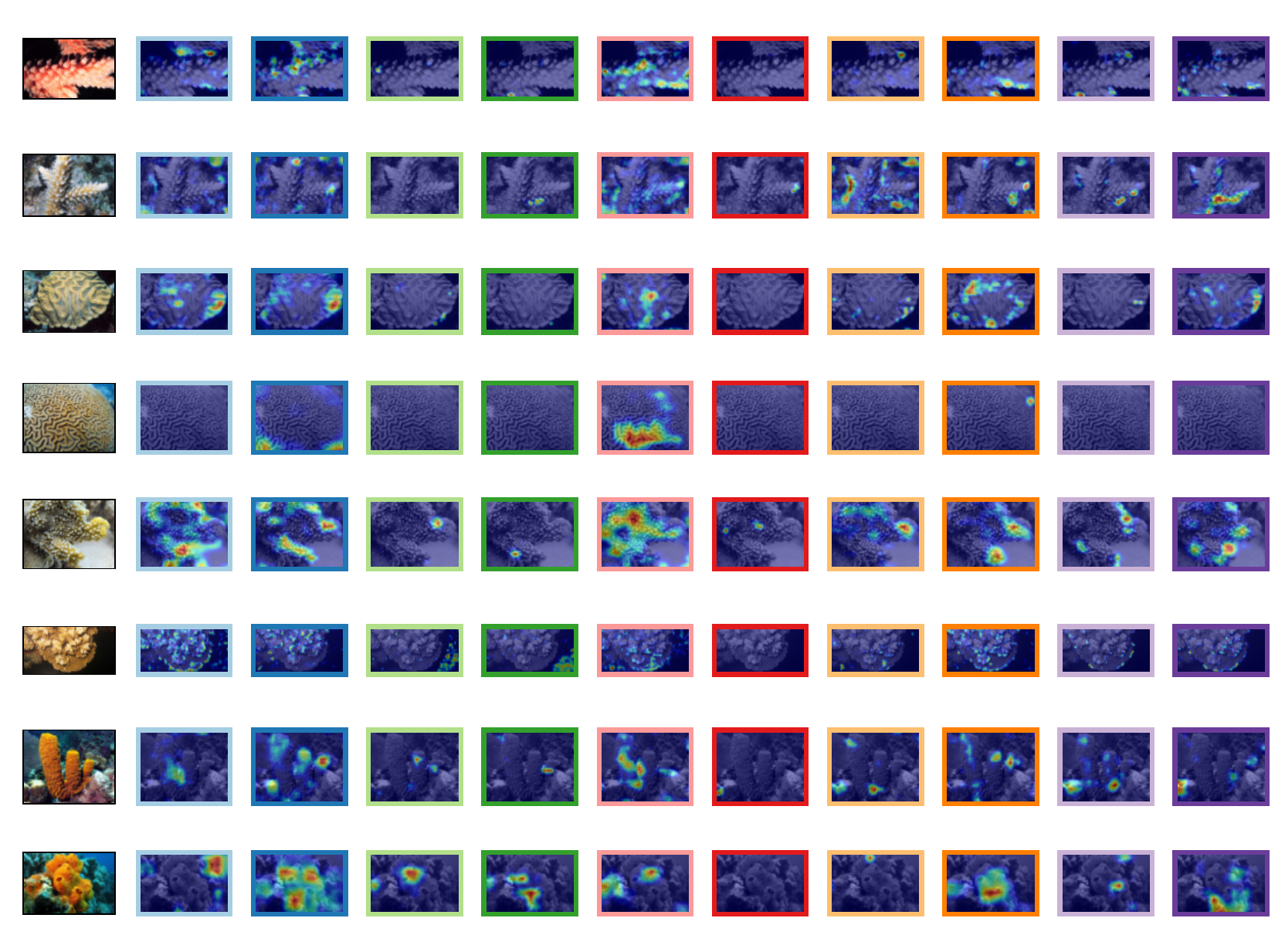}}
    \caption{Bars, boxplots and heatmaps of SRSMAS.}
    \label{fig:xaisrsmas}
\end{figure}

The last dataset is SRSMAS, whose charts are presented in Figure \ref{fig:xaisrsmas}. The most relevant neurons obtain a minimum of 60\% of appearance in all the solutions of the Pareto front, which has been a constant factor in all the datasets. Moreover, the distribution of the objectives for the solutions, in which these neurons are active, shares a common line: high values both performance and robustness and low complexity of the network. These neurons draw pruning patterns that identify the class for the input images. As an example, in the fourth image it is only necessary to recognize the silhouette of the coral reef, but in the fifth one, the network needs to understand how is the central part of the coral reef and then its extremities.

In these figures, we have seen several datasets in which the difference rate between the most important neurons is close (RPS and PLANTS), but there are other datasets in which this difference is up to 20\% between the most and least relevant neurons. Moreover, the distribution of the objectives for each dataset gives us good insights about the uniformity of the performance and robustness in most of the datasets.

The good work done by \proposal in training the models has made possible to achieve remarkable pruning patterns. These have helped us to decipher not only those neurons that have been key in the whole training and inference process, but also to locate in the input images those groups of influential pixels which have been important to decide the class of each of these images.

\subsection{Answering RQ3: Quality of the models through ensemble modeling} \label{sec:rq3}

\begin{idea}

    This subsection is devised to formally answer to RQ3, which is to show if the ensemble modeling is able to improve the quality of the trained models by \proposal. An elementary key in this regard is model diversity, understood as the ability to generate and train models from a given dataset that are different to each other and that model differently the distribution underlying the dataset at hand. If \proposal is found to be capable of generating models in this way (as a byproduct of its multi-objective search), then an ensemble of such models can give rise to an improved performance and reduced risk of overfitting. For all of these reasons, ensemble modeling can achieve an improvement in terms of either accuracy and/or robustness with respect to the individual pruned models comprised in the Pareto front estimated by \proposal.
    
    Having established the motivation for ensemble modeling, we will now describe its implementation. We depart from the two objectives to be maximized, namely, accuracy and robustness. The proposed ensemble strategy consists of collecting the models in the estimated Pareto front (containing the best solutions from the different runs performed) that fall within a statistical range of accuracy or AUROC (the robustness measure). Thus, the ensemble will fuse together those pruned models that fall within two given percentiles of the distribution of these metrics over the Pareto front of the three objectives. From these assembled models, their predictions for a given query are merged into one (by simple majority voting), and compared to the prediction of the best individual model in the ensemble.

    The analysis of the ensemble behavior is done based on different percentile ranges of each of the accuracy and AUROC distributions, providing more precise information for each of these metrics. Next, we explain how such percentile ranges are chosen. We start with a first range (percentiles $(50\%, 60\%)$), and we increase the extremes of the interval by 5\% in each iteration, giving us 8 quartile intervals for both metrics. Models in the Pareto front whose objective values fall within each of these percentile ranges are included in the ensemble. For example, the interval $(75\%,85\%)$ will contain those models in the estimated Pareto front whose accuracy objective is within this range given the distribution of the accuracy objective computed over the whole Pareto front estimation (a similar example can be given for the AUROC score). These percentiles are defined as $(Q_{min},Q_{max})$, where $Q_{min} = 50\%, 55\%, \dots, 85\%$, and $Q_{max} = 60\%, 65\%, \dots, 95\%$. With this division, we have the following intervals $(50\%,60\%), (55\%,65\%), \dots, (85\%,95\%)$.
\end{idea}

\begin{figure}[!hbtp]
    \includegraphics[width=\linewidth]{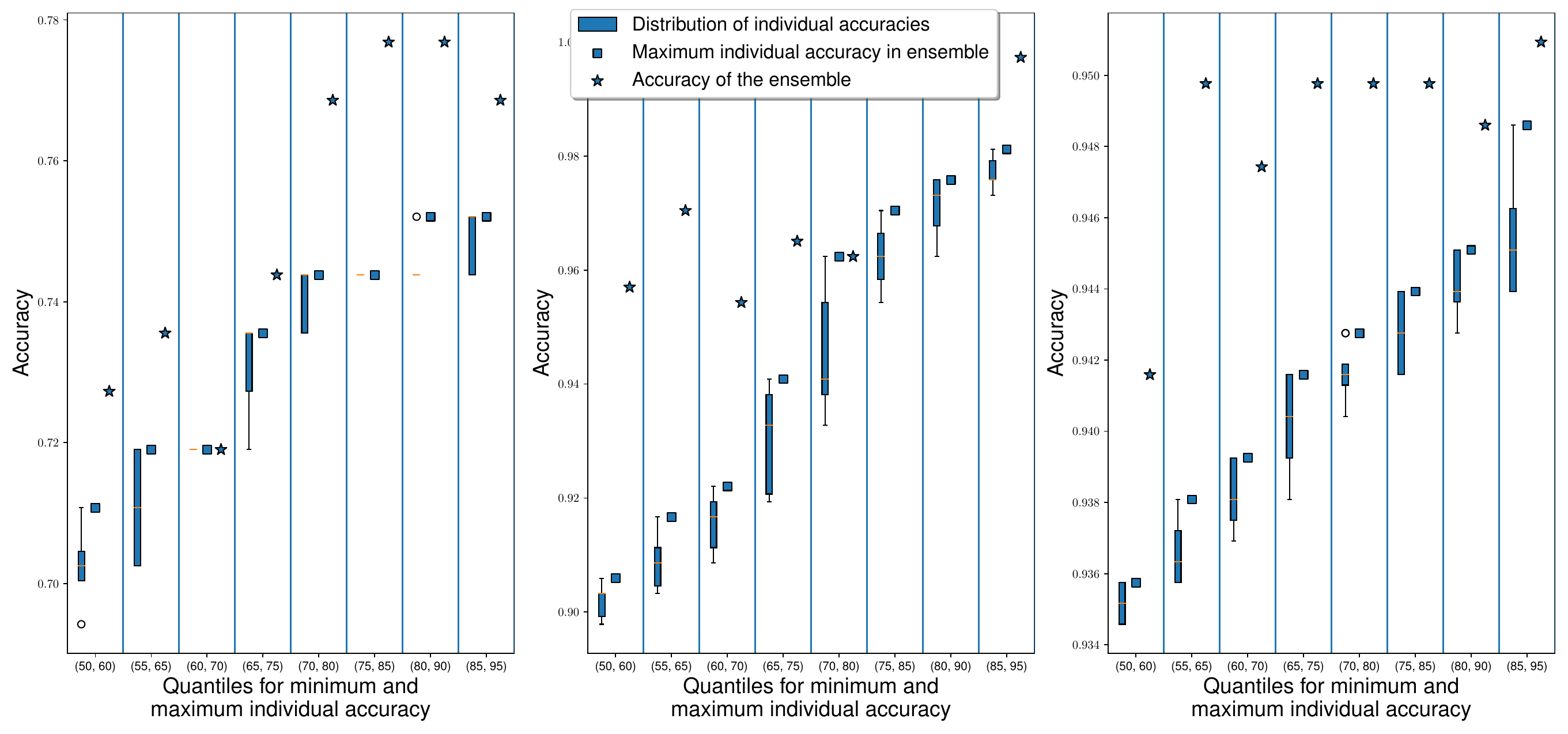}
    \caption{Ensemble modeling of the models trained by \proposal in terms of accuracy. Left: CATARACT dataset. Middle: RPS dataset. Right: PAINTING dataset.}
    \label{fig:ensembleAcc}
\end{figure}

\begin{idea}
    In this study, we have selected the CATARACT, PAINTING and RPS datasets for the experimental tests performed to examine the behavior of ensemble modeling. Two different plots are depicted for each dataset, one for each metric (accuracy and AUROC). In each of these plots, three symbols appear in the form of a rectangle, a square and a star. The rectangle shows the distribution of accuracy/AUROC values for the models in the percentile range at hand. The square symbolizes the best result for that measure. Lastly, the star indicates the accuracy/AUROC of the ensemble.\end{idea}With this explanation, we can interpret the two graphs that result from making the ensemble. The first one is related to the accuracy of the network. Figure \ref{fig:ensembleAcc} shows this graph. It presents three graphs sorted alphabetically by dataset. The first one corresponds to CATARACT, the second to RPS and the third to PAINTING. Each of them shows, for each interval of quantile the distribution of individual accuracies, the maximum of the distribution and the accuracy of the ensemble.

The overall performance in the three cases is positive since the diversity of the models allows us to find new models that improve the accuracy for each quantile interval, except in the case of RPS where we only have one model in the interval $(60\%,70\%)$. In the RPS case, we have models near the 90\% of accuracy and the ensemble produces a new model with almost 96\% of it, which is a great result. Moreover, models with higher accuracy (96\% or more) achieve close to 100\% of accuracy. For RPS (the chart of the right), most of the ensemble models get a 95\% of accuracy, meanwhile their individual models are present a lower value in accuracy. As a result, these charts show the benefits of the ensemble modeling for the accuracy objective.

The second part of this section consists of replicating the previous experiment, but for the case of OoD detection in order to check if the AUROC improves when ensemble modeling occurs. In such a case, we will be able to confirm that the new model detects less OoD sample as InD, which makes an improvement in the associated metric.

\begin{figure}[!hbtp]
    \includegraphics[width=\linewidth]{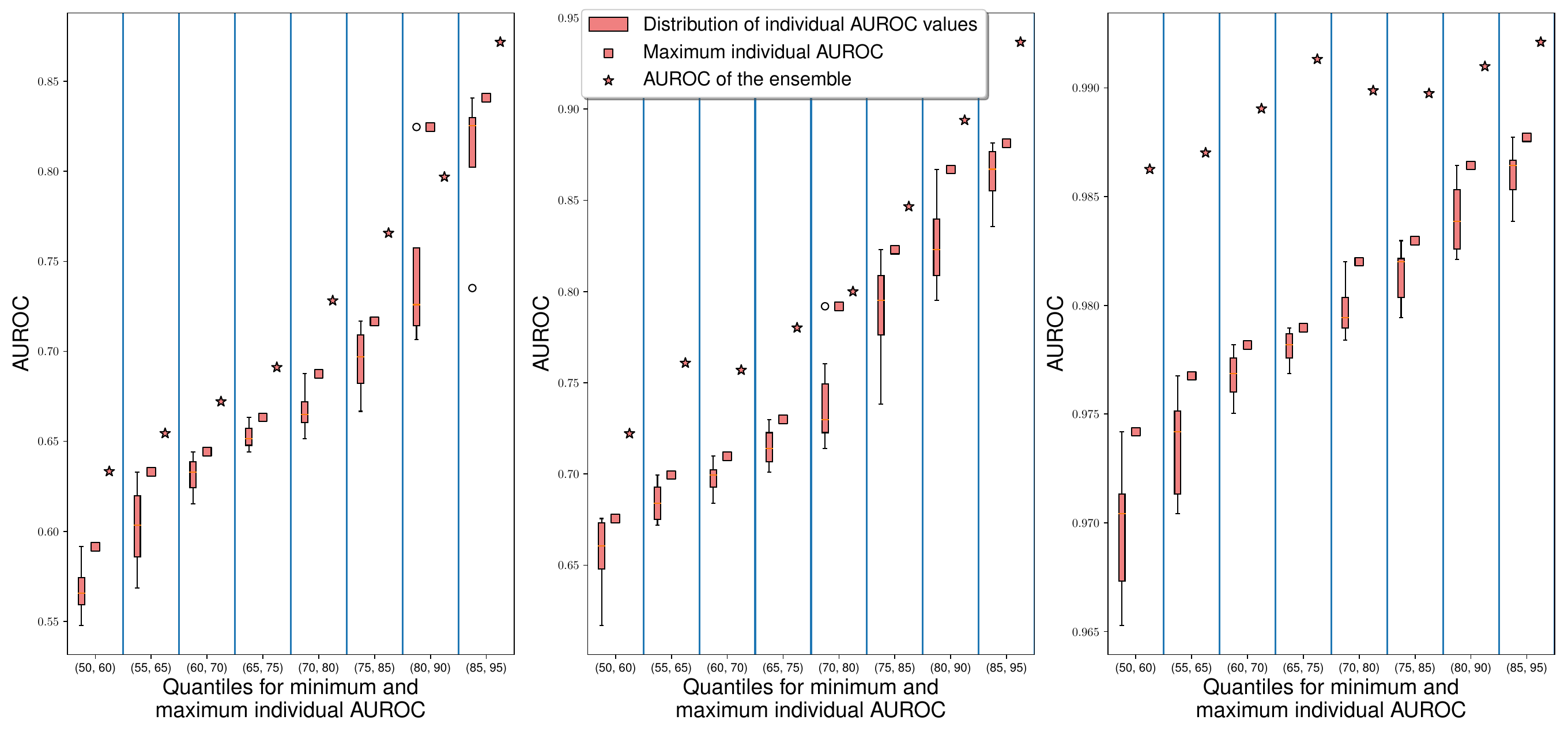}
    \caption{Ensemble modeling of the models trained by \proposal in terms of OoD detection. Left: CATARACT dataset. Middle: RPS dataset. Right: PAINTING dataset.}
    \label{fig:ensembleOoD}
\end{figure}

The interpretation of the set of charts is the same as in the previous case. We have made the ensemble with the models for each interval. The same characteristics are presented in Fig. \ref{fig:ensembleOoD}. It is shown the distribution of individual AUROC values, its maximum and then, marked with a star, the AUROC of the ensemble. The CATARACT case shows an improvement in the AUROC in all the cases but one. For the RPS case (middle chart), the case of $(85\%,95\%)$ achieves almost a 95\% of AUROC, meanwhile the individual values get a maximum of 87\%. The PAINTING dataset also presents outstanding results. Its minimum AUROC for all the intervals of the ensemble is more than 98.5\% and the least value is of individual models is less than 97\%. The results obtained from the graph are similar to those obtained for the case of accuracy, since they improve on the individual results in the vast majority of the intervals.

In this section, we have conducted two experiments which involve the ensemble modeling of the trained models by \proposal. The ensemble has been done taking into account the performance of the network and the robustness and we have given the liberty to choose the interval of values for each measure. The results drawn from these graphics show that both of the objectives have been improved. Performing a MO search not only provides the user with a wide range of models that balance between the three stated objectives, but it also achieves more diversity among the models in order to ensemble them and achieve even higher performance and robustness.

\section{Conclusions} \label{sec:conclusions}

This paper has introduced \proposal, a MONAS model that evolves sparse layers of a DL model which has been instantiated using the TL paradigm. \proposal uses a MOEA, which evolves these sparse layers, in order to obtain adapted, pruned layers to the problem at hand and making decisions about the neurons that need to be active or inactive. 

\proposal is a model that evolves the extracted features from the pre-trained network in order to train the last layers to tackle the considered problem. Our results draw two conclusions from the Pareto fronts: there exists a great diversity in the solutions and they also establish promising values for the objectives in their extremes values. Moreover, the projections for each objective shed light on the existence of a direct relationships between the complexity of the network and each of the other two objectives (performance and robustness), whereas there is no direct relationship between the latter two. This work falls within the umbrella of OWL because the evolved models are asked about new data, which is the OoD datasets. Moreover, OWL is related with GPAI and, particularly, in this manuscript, the experiments have shown the capability of AI generating AI as the MOEA has learnt from the trained DL models.

The trained models of \proposal lead to several pruning patterns in which there exist neurons that appear in most of the best solutions of the Pareto front. These patterns help us to recognize the key group of regions of the input images that our models consider the most important ones when assigning the class to the input image at inference time.

The diversity of the models of \proposal has shown that ensemble modeling is able to increase the overall performance, both in performance of the network and robustness, in most of the quantiles for minimum and maximum considered objective values.

The evolved trained models have shown a great performance with a minimum number of active neurons, but it is also shown the great contribution of the robustness for these models, as each DL model is tested with data that it has not previously seen. Moreover, the objectives of the MOEA have been the performance, complexity and robustness, but other alternatives can be formulated as objectives such as the latency or energy used of the GPU in the inference of the pruned model or the epistemic uncertainty level.

An ablation study is also in our agenda for future research, aiming to discern which algorithmic steps are more relevant for the search convergence of the solver when tackling the multi-objective problem at hand. We envision that the results of this ablation study can illuminate the design of new operators and more effective search strategies than the ones utilized in this work. \begin{idea}Moreover, we will investigate the influence of different robustness measures on the Pareto front estimations produced by \proposal. \end{idea}

\section*{Acknowledgments}

F. Herrera, D. Molina and J. Poyatos are supported by the Andalusian Excellence project P18-FR-4961, the infrastructure project with reference EQC2018-005084-P and the R\&D and Innovation project with reference PID2020-119478GB-I00 granted by the Spain's Ministry of Science and Innovation and European Regional Development Fund (ERDF). A. Martinez-Seras and J. Del Ser would like to thank the Basque Government for the funding support received through the EMAITEK and ELKARTEK programs, as well as the Consolidated Research Group MATHMODE (IT1456-22) granted by the Department of Education of this institution.

\bibliographystyle{model5-names}


\bibliography{bibliography}

\begin{thebibliography}{62}
\expandafter\ifx\csname natexlab\endcsname\relax\def\natexlab#1{#1}\fi
\providecommand{\url}[1]{\texttt{#1}}
\providecommand{\href}[2]{#2}
\providecommand{\path}[1]{#1}
\providecommand{\DOIprefix}{doi:}
\providecommand{\ArXivprefix}{arXiv:}
\providecommand{\URLprefix}{URL: }
\providecommand{\Pubmedprefix}{pmid:}
\providecommand{\doi}[1]{\href{http://dx.doi.org/#1}{\path{#1}}}
\providecommand{\Pubmed}[1]{\href{pmid:#1}{\path{#1}}}
\providecommand{\bibinfo}[2]{#2}
\ifx\xfnm\relax \def\xfnm[#1]{\unskip,\space#1}\fi
\bibitem[{Assun{\c{c}}{\~a}o et~al.(2019)Assun{\c{c}}{\~a}o, Louren{\c{c}}o,
  Machado \& Ribeiro}]{assunccao2019denser}
\bibinfo{author}{Assun{\c{c}}{\~a}o, F.}, \bibinfo{author}{Louren{\c{c}}o, N.},
  \bibinfo{author}{Machado, P.}, \& \bibinfo{author}{Ribeiro, B.}
  (\bibinfo{year}{2019}).
\newblock \bibinfo{title}{Denser: deep evolutionary network structured
  representation}.
\newblock {\it \bibinfo{journal}{Genetic Programming and Evolvable
  Machines}\/},  {\it \bibinfo{volume}{20}\/}, \bibinfo{pages}{5--35}.
\newblock \bibinfo{note}{\url{https://doi.org/10.1007/s10710-018-9339-y}}.
\bibitem[{Back et~al.(1997)Back, Fogel \& Michalewicz}]{EABook}
\bibinfo{author}{Back, T.}, \bibinfo{author}{Fogel, D.~B.}, \&
  \bibinfo{author}{Michalewicz, Z.} (\bibinfo{year}{1997}).
\newblock {\it \bibinfo{title}{{Handbook of Evolutionary Computation}}\/}.
\newblock (\bibinfo{edition}{1st} ed.).
\newblock \bibinfo{publisher}{IOP Publishing Ltd.}
\bibitem[{Back \& Schwefel(1996, May)}]{evocomp}
\bibinfo{author}{Back, T.}, \& \bibinfo{author}{Schwefel, H.-P.}
  (\bibinfo{year}{1996, May}).
\newblock \bibinfo{title}{{Evolutionary computation: an overview}}.
\newblock In {\it \bibinfo{booktitle}{Proceedings of IEEE International
  Conference on Evolutionary Computation, Padua, Italy, 1996}\/}.
\newblock \bibinfo{note}{\url{https://doi.org/10.1109/ICEC.1996.542329}}.
\bibitem[{{Baldeon Calisto} \& Lai-Yuen(2020)}]{baldeon}
\bibinfo{author}{{Baldeon Calisto}, M.}, \& \bibinfo{author}{Lai-Yuen, S.~K.}
  (\bibinfo{year}{2020}).
\newblock \bibinfo{title}{{AdaEn-Net: An ensemble of adaptive 2D–3D Fully
  Convolutional Networks for medical image segmentation}}.
\newblock {\it \bibinfo{journal}{Neural Networks}\/},  {\it
  \bibinfo{volume}{126}\/}, \bibinfo{pages}{76--94}.
\newblock \bibinfo{note}{\url{https://doi.org/10.1016/j.neunet.2020.03.007}}.
\bibitem[{Baldeon-Calisto \& Lai-Yuen(2020)}]{BALDEONCALISTO2020325}
\bibinfo{author}{Baldeon-Calisto, M.}, \& \bibinfo{author}{Lai-Yuen, S.~K.}
  (\bibinfo{year}{2020}).
\newblock \bibinfo{title}{{AdaResU-Net: Multiobjective adaptive convolutional
  neural network for medical image segmentation}}.
\newblock {\it \bibinfo{journal}{Neurocomputing}\/},  {\it
  \bibinfo{volume}{392}\/}, \bibinfo{pages}{325--340}.
\newblock \bibinfo{note}{\url{https://doi.org/10.1016/j.neucom.2019.01.110}}.
\bibitem[{{Barredo Arrieta} et~al.(2020){Barredo Arrieta}, Díaz-Rodríguez,
  {Del Ser}, Bennetot, Tabik, Barbado, Garcia, Gil-Lopez, Molina, Benjamins,
  Chatila \& Herrera}]{xaibarredo}
\bibinfo{author}{{Barredo Arrieta}, A.}, \bibinfo{author}{Díaz-Rodríguez,
  N.}, \bibinfo{author}{{Del Ser}, J.}, \bibinfo{author}{Bennetot, A.},
  \bibinfo{author}{Tabik, S.}, \bibinfo{author}{Barbado, A.},
  \bibinfo{author}{Garcia, S.}, \bibinfo{author}{Gil-Lopez, S.},
  \bibinfo{author}{Molina, D.}, \bibinfo{author}{Benjamins, R.},
  \bibinfo{author}{Chatila, R.}, \& \bibinfo{author}{Herrera, F.}
  (\bibinfo{year}{2020}).
\newblock \bibinfo{title}{{Explainable Artificial Intelligence (XAI): Concepts,
  taxonomies, opportunities and challenges toward responsible AI}}.
\newblock {\it \bibinfo{journal}{Information Fusion}\/},  {\it
  \bibinfo{volume}{58}\/}, \bibinfo{pages}{82--115}.
\newblock \bibinfo{note}{\url{https://doi.org/10.1016/j.inffus.2019.12.012}}.
\bibitem[{Calisto \& Lai-Yuen(2020)}]{calistonas}
\bibinfo{author}{Calisto, M.~B.}, \& \bibinfo{author}{Lai-Yuen, S.}
  (\bibinfo{year}{2020}).
\newblock \bibinfo{title}{{Neural Architecture Search with an Efficient
  Multiobjective Evolutionary Framework}}.
\newblock {\it \bibinfo{journal}{arXiv preprint arXiv:2011.04463}\/}, .
\newblock \bibinfo{note}{\url{https://arxiv.org/abs/2011.04463}}.
\bibitem[{Chitty-Venkata \& Somani(2022)}]{nashardware}
\bibinfo{author}{Chitty-Venkata, K.~T.}, \& \bibinfo{author}{Somani, A.~K.}
  (\bibinfo{year}{2022}).
\newblock \bibinfo{title}{{Neural Architecture Search Survey: A Hardware
  Perspective}}.
\newblock {\it \bibinfo{journal}{ACM Computing Surveys}\/},  {\it
  \bibinfo{volume}{55}\/}, \bibinfo{pages}{1--36}.
\newblock \bibinfo{note}{\url{https://doi.org/10.1145/3524500}}.
\bibitem[{[dataset]Sungjoon Choi(2020)}]{Eyes}
\bibinfo{author}{[dataset]Sungjoon Choi} (\bibinfo{year}{2020}).
\newblock \bibinfo{title}{{Cataract Dataset}}.
\newblock \bibinfo{note}{Retrieved from
  \url{https://www.kaggle.com/jr2ngb/cataractdataset}. Accessed September 10,
  2020}.
\bibitem[{Clune(2019)}]{aigenai}
\bibinfo{author}{Clune, J.} (\bibinfo{year}{2019}).
\newblock \bibinfo{title}{{AI-GAs: AI-generating algorithms, an alternate
  paradigm for producing general artificial intelligence}}.
\newblock {\it \bibinfo{journal}{arXiv prepint arXiv:1905.10985}\/}, .
\newblock \bibinfo{note}{\url{https://arxiv.org/abs/1905.10985}}.
\bibitem[{Deb(2011)}]{Deb2011}
\bibinfo{author}{Deb, K.} (\bibinfo{year}{2011}).
\newblock {\it \bibinfo{title}{Multi-objective Optimisation Using Evolutionary
  Algorithms: An Introduction}\/}.
\newblock \bibinfo{publisher}{Springer London}.
\bibitem[{Deb et~al.(2002)Deb, Pratap, Agarwal \& Meyarivan}]{nsga}
\bibinfo{author}{Deb, K.}, \bibinfo{author}{Pratap, A.},
  \bibinfo{author}{Agarwal, S.}, \& \bibinfo{author}{Meyarivan, T.}
  (\bibinfo{year}{2002}).
\newblock \bibinfo{title}{{A fast and elitist multiobjective genetic algorithm:
  NSGA-II}}.
\newblock {\it \bibinfo{journal}{IEEE Transactions on Evolutionary
  Computation}\/},  {\it \bibinfo{volume}{6}\/}, \bibinfo{pages}{182--197}.
\newblock \bibinfo{note}{\url{https://doi.org/10.1109/4235.996017}}.
\bibitem[{Dufourq \& Bassett(2017, November)}]{dufourq}
\bibinfo{author}{Dufourq, E.}, \& \bibinfo{author}{Bassett, B.~A.}
  (\bibinfo{year}{2017, November}).
\newblock \bibinfo{title}{{EDEN: Evolutionary deep networks for efficient
  machine learning}}.
\newblock In {\it \bibinfo{booktitle}{2017 Pattern Recognition Association of
  South Africa and Robotics and Mechatronics (PRASA-RobMech), Bloemfontein,
  South Africa, 2017}\/} (pp. \bibinfo{pages}{110--115}).
\newblock \bibinfo{note}{\url{https://doi.org/10.1109/RoboMech.2017.8261132}}.
\bibitem[{Elsken et~al.(2019)Elsken, Metzen \& Hutter}]{elsken2019neural}
\bibinfo{author}{Elsken, T.}, \bibinfo{author}{Metzen, J.~H.}, \&
  \bibinfo{author}{Hutter, F.} (\bibinfo{year}{2019}).
\newblock \bibinfo{title}{{Neural architecture search: A survey}}.
\newblock {\it \bibinfo{journal}{The Journal of Machine Learning Research}\/},
  {\it \bibinfo{volume}{20}\/}, \bibinfo{pages}{1997--2017}.
\newblock \bibinfo{note}{\url{http://jmlr.org/papers/v20/18-598.html}}.
\bibitem[{Elsken et~al.(2019, May)Elsken, Metzen \&
  Hutter}]{Elsken2019EfficientMN}
\bibinfo{author}{Elsken, T.}, \bibinfo{author}{Metzen, J.~H.}, \&
  \bibinfo{author}{Hutter, F.} (\bibinfo{year}{2019, May}).
\newblock \bibinfo{title}{{Efficient Multi-Objective Neural Architecture Search
  via Lamarckian Evolution}}.
\newblock In {\it \bibinfo{booktitle}{7th International Conference on Learning
  Representations (ICLR), New Orleans, LA, USA, 2019}\/}.
\newblock \bibinfo{note}{\url{https://doi.org/10.48550/arXiv.1804.09081}}.
\bibitem[{[dataset] G\'omez-R\'ios et~al.(2019)[dataset] G\'omez-R\'ios, Tabik,
  Luengo, Shihavuddin \& Herrera}]{SRSMAS}
\bibinfo{author}{[dataset] G\'omez-R\'ios, A.}, \bibinfo{author}{Tabik, S.},
  \bibinfo{author}{Luengo, J.}, \bibinfo{author}{Shihavuddin, A.}, \&
  \bibinfo{author}{Herrera, F.} (\bibinfo{year}{2019}).
\newblock \bibinfo{title}{{Coral species identification with texture or
  structure images using a two-level classifier based on Convolutional Neural
  Networks}}.
\newblock {\it \bibinfo{journal}{Knowledge-Based Systems}\/},  {\it
  \bibinfo{volume}{184}\/}, \bibinfo{pages}{104891}.
\newblock \bibinfo{note}{\url{https://doi.org/10.1016/j.knosys.2019.104891}}.
\bibitem[{Han et~al.(2015, December)Han, Pool, Tran \& Dally}]{han2015learning}
\bibinfo{author}{Han, S.}, \bibinfo{author}{Pool, J.}, \bibinfo{author}{Tran,
  J.}, \& \bibinfo{author}{Dally, W.} (\bibinfo{year}{2015, December}).
\newblock \bibinfo{title}{{Learning both Weights and Connections for Efficient
  Neural Network}}.
\newblock In {\it \bibinfo{booktitle}{Proceedings of the 28th International
  Conference on Neural Information Processing Systems, Montreal Canada,
  2015}\/}.
\newblock
  \bibinfo{note}{\url{https://papers.nips.cc/paper_files/paper/2015/hash/ae0eb3eed39d2bcef4622b2499a05fe6-Abstract.html}}.
\bibitem[{Hendrycks \& Gimpel(2016)}]{hendrycks2016baseline}
\bibinfo{author}{Hendrycks, D.}, \& \bibinfo{author}{Gimpel, K.}
  (\bibinfo{year}{2016}).
\newblock \bibinfo{title}{{A baseline for detecting misclassified and
  out-of-distribution examples in neural networks}}.
\newblock {\it \bibinfo{journal}{arXiv preprint arXiv:1610.02136}\/}, .
\newblock \bibinfo{note}{\url{https://doi.org/10.48550/arXiv.1610.02136}}.
\bibitem[{Hendrycks et~al.(2019, May)Hendrycks, Mazeika \&
  Dietterich}]{hendrycks2018deep}
\bibinfo{author}{Hendrycks, D.}, \bibinfo{author}{Mazeika, M.}, \&
  \bibinfo{author}{Dietterich, T.} (\bibinfo{year}{2019, May}).
\newblock \bibinfo{title}{Deep anomaly detection with outlier exposure}.
\newblock In {\it \bibinfo{booktitle}{International Conference on Learning
  Representations, New Orleans, USA, 2019}\/}.
\newblock \bibinfo{note}{\url{https://openreview.net/forum?id=HyxCxhRcY7}}.
\bibitem[{Hoefler et~al.(2021)Hoefler, Alistarh, Ben-Nun, Dryden \&
  Peste}]{Hoefler2021}
\bibinfo{author}{Hoefler, T.}, \bibinfo{author}{Alistarh, D.},
  \bibinfo{author}{Ben-Nun, T.}, \bibinfo{author}{Dryden, N.}, \&
  \bibinfo{author}{Peste, A.} (\bibinfo{year}{2021}).
\newblock \bibinfo{title}{{Sparsity in Deep Learning: Pruning and Growth for
  Efficient Inference and Training in Neural Networks}}.
\newblock {\it \bibinfo{journal}{Journal of Machine Learning Research}\/},
  {\it \bibinfo{volume}{22}\/}, \bibinfo{pages}{1--124}.
\bibitem[{ISO(2021{\natexlab{a}})}]{iso1}
\bibinfo{author}{ISO} (\bibinfo{year}{2021}{\natexlab{a}}).
\newblock {\it \bibinfo{title}{{Artificial Intelligence (AI) — Assessment of
  the robustness of neural networks — Part 1: Overview}}\/}.
\newblock \bibinfo{type}{Technical Report} ISO/IEC JTC 1/SC 42 Artificial
  intelligence (24029-1:2021).
\bibitem[{ISO(2021{\natexlab{b}})}]{iso2}
\bibinfo{author}{ISO} (\bibinfo{year}{2021}{\natexlab{b}}).
\newblock {\it \bibinfo{title}{{Artificial intelligence (AI) — Assessment of
  the robustness of neural networks — Part 2: Methodology for the use of
  formal methods}}\/}.
\newblock \bibinfo{type}{Technical Report} ISO/IEC JTC 1/SC 42 Artificial
  intelligence (24029-2).
\bibitem[{Khan et~al.(2019)Khan, Islam, Jan, Din \& Rodrigues}]{khan2019novel}
\bibinfo{author}{Khan, S.}, \bibinfo{author}{Islam, N.}, \bibinfo{author}{Jan,
  Z.}, \bibinfo{author}{Din, I.~U.}, \& \bibinfo{author}{Rodrigues, J. J.~C.}
  (\bibinfo{year}{2019}).
\newblock \bibinfo{title}{{A novel deep learning based framework for the
  detection and classification of breast cancer using transfer learning}}.
\newblock {\it \bibinfo{journal}{Pattern Recognition Letters}\/},  {\it
  \bibinfo{volume}{125}\/}, \bibinfo{pages}{1--6}.
\newblock \bibinfo{note}{\url{https://doi.org/10.1016/j.patrec.2019.03.022}}.
\bibitem[{Krizhevsky et~al.(2017)Krizhevsky, Sutskever \& Hinton}]{imagenet}
\bibinfo{author}{Krizhevsky, A.}, \bibinfo{author}{Sutskever, I.}, \&
  \bibinfo{author}{Hinton, G.~E.} (\bibinfo{year}{2017}).
\newblock \bibinfo{title}{{ImageNet Classification with Deep Convolutional
  Neural Networks}}.
\newblock {\it \bibinfo{journal}{Commun. ACM}\/},  {\it
  \bibinfo{volume}{60}\/}, \bibinfo{pages}{84–90}.
\newblock \bibinfo{note}{\url{https://doi.org/10.1145/3065386}}.
\bibitem[{Lee et~al.(2018)Lee, Lee, Lee \& Shin}]{lee2018simple}
\bibinfo{author}{Lee, K.}, \bibinfo{author}{Lee, K.}, \bibinfo{author}{Lee,
  H.}, \& \bibinfo{author}{Shin, J.} (\bibinfo{year}{2018}).
\newblock \bibinfo{title}{{A simple unified framework for detecting
  out-of-distribution samples and adversarial attacks}}.
\newblock {\it \bibinfo{journal}{Advances in Neural Information Processing
  Systems}\/},  {\it \bibinfo{volume}{31}\/}.
\newblock
  \bibinfo{note}{\url{https://proceedings.neurips.cc/paper/2018/file/abdeb6f575ac5c6676b747bca8d09cc2-Paper.pdf}}.
\bibitem[{Liang et~al.(2018, April{\natexlab{a}})Liang, Li \&
  Srikant}]{liang2017odin}
\bibinfo{author}{Liang, S.}, \bibinfo{author}{Li, Y.}, \&
  \bibinfo{author}{Srikant, R.} (\bibinfo{year}{2018, April}{\natexlab{a}}).
\newblock \bibinfo{title}{{Enhancing the reliability of out-of-distribution
  image detection in neural networks}}.
\newblock In {\it \bibinfo{booktitle}{Proceedings of the 6th International
  Conference on Learning Representations (ICLR), Vancouver, Canada, 2018}\/}.
\newblock \bibinfo{note}{\url{https://doi.org/10.48550/arXiv.1706.02690}}.
\bibitem[{Liang et~al.(2018, April{\natexlab{b}})Liang, Li \&
  Srikant}]{liang2018enhancing}
\bibinfo{author}{Liang, S.}, \bibinfo{author}{Li, Y.}, \&
  \bibinfo{author}{Srikant, R.} (\bibinfo{year}{2018, April}{\natexlab{b}}).
\newblock \bibinfo{title}{{Enhancing The Reliability of Out-of-distribution
  Image Detection in Neural Networks}}.
\newblock In {\it \bibinfo{booktitle}{International Conference on Learning
  Representations, Vancouver, Canada, 2018}\/}.
\newblock \bibinfo{note}{\url{https://openreview.net/forum?id=H1VGkIxRZ}}.
\bibitem[{Lin et~al.(2021, June)Lin, Roy \& Li}]{lin2021mood}
\bibinfo{author}{Lin, Z.}, \bibinfo{author}{Roy, S.~D.}, \&
  \bibinfo{author}{Li, Y.} (\bibinfo{year}{2021, June}).
\newblock \bibinfo{title}{{{MOOD}: Multi-level Out-of-distribution Detection}}.
\newblock In {\it \bibinfo{booktitle}{2021 {IEEE}/{CVF} Conference on Computer
  Vision and Pattern Recognition ({CVPR}), Nashville, Tennessee ,2021}\/}.
\newblock \bibinfo{publisher}{{IEEE}}.
\newblock \bibinfo{note}{\url{https://doi.org/10.1109/cvpr46437.2021.01506}}.
\bibitem[{Liu et~al.(2020)Liu, Wang, Owens \& Li}]{liu2020energy}
\bibinfo{author}{Liu, W.}, \bibinfo{author}{Wang, X.}, \bibinfo{author}{Owens,
  J.}, \& \bibinfo{author}{Li, Y.} (\bibinfo{year}{2020}).
\newblock \bibinfo{title}{{Energy-based out-of-distribution detection}}.
\newblock {\it \bibinfo{journal}{Advances in Neural Information Processing
  Systems}\/},  {\it \bibinfo{volume}{33}\/}, \bibinfo{pages}{21464--21475}.
\newblock
  \bibinfo{note}{\url{https://proceedings.neurips.cc/paper/2020/hash/f5496252609c43eb8a3d147ab9b9c006-Abstract.html}}.
\bibitem[{Loni et~al.(2020)Loni, Sinaei, Zoljodi, Daneshtalab \&
  Sjödin}]{LONI2020102989}
\bibinfo{author}{Loni, M.}, \bibinfo{author}{Sinaei, S.},
  \bibinfo{author}{Zoljodi, A.}, \bibinfo{author}{Daneshtalab, M.}, \&
  \bibinfo{author}{Sjödin, M.} (\bibinfo{year}{2020}).
\newblock \bibinfo{title}{{DeepMaker: A multi-objective optimization framework
  for deep neural networks in embedded systems}}.
\newblock {\it \bibinfo{journal}{Microprocessors and Microsystems}\/},  {\it
  \bibinfo{volume}{73}\/}, \bibinfo{pages}{102989}.
\newblock \bibinfo{note}{\url{https://doi.org/10.1016/j.micpro.2020.102989}}.
\bibitem[{Lu et~al.(2022{\natexlab{a}})Lu, Cheng, Huang, Zhang, Qiu \&
  Yang}]{surrogate}
\bibinfo{author}{Lu, Z.}, \bibinfo{author}{Cheng, R.}, \bibinfo{author}{Huang,
  S.}, \bibinfo{author}{Zhang, H.}, \bibinfo{author}{Qiu, C.}, \&
  \bibinfo{author}{Yang, F.} (\bibinfo{year}{2022}{\natexlab{a}}).
\newblock \bibinfo{title}{{Surrogate-assisted Multi-objective Neural
  Architecture Search for Real-time Semantic Segmentation}}.
\newblock {\it \bibinfo{journal}{arXiv preprint arXiv:2208.06820}\/}, .
\newblock \bibinfo{note}{\url{https://arxiv.org/abs/2208.06820}}.
\bibitem[{Lu et~al.(2022{\natexlab{b}})Lu, Cheng, Jin, Tan \&
  Deb}]{lu2022neural}
\bibinfo{author}{Lu, Z.}, \bibinfo{author}{Cheng, R.}, \bibinfo{author}{Jin,
  Y.}, \bibinfo{author}{Tan, K.~C.}, \& \bibinfo{author}{Deb, K.}
  (\bibinfo{year}{2022}{\natexlab{b}}).
\newblock \bibinfo{title}{{Neural Architecture Search as Multiobjective
  Optimization Benchmarks: Problem Formulation and Performance Assessment}}.
\newblock {\it \bibinfo{journal}{arXiv preprint arXiv:2208.04321}\/}, .
\newblock \bibinfo{note}{\url{https://doi.org/10.48550/arXiv.2208.04321}}.
\bibitem[{Lu et~al.(2020, August)Lu, Deb, Goodman, Banzhaf \&
  Boddeti}]{nsganetv2}
\bibinfo{author}{Lu, Z.}, \bibinfo{author}{Deb, K.}, \bibinfo{author}{Goodman,
  E.}, \bibinfo{author}{Banzhaf, W.}, \& \bibinfo{author}{Boddeti, V.~N.}
  (\bibinfo{year}{2020, August}).
\newblock \bibinfo{title}{{NSGANetV2: Evolutionary Multi-Objective
  Surrogate-Assisted Neural Architecture Search}}.
\newblock In {\it \bibinfo{booktitle}{Computer Vision – ECCV 2020: 16th
  European Conference, Glasgow, UK, Proceedings, Part I}\/}.
\newblock \bibinfo{note}{\url{https://doi.org/10.1007/978-3-030-58452-8_3}}.
\bibitem[{Lu et~al.(2019, July)Lu, Whalen, Boddeti, Dhebar, Deb, Goodman \&
  Banzhaf}]{nsganet}
\bibinfo{author}{Lu, Z.}, \bibinfo{author}{Whalen, I.},
  \bibinfo{author}{Boddeti, V.}, \bibinfo{author}{Dhebar, Y.},
  \bibinfo{author}{Deb, K.}, \bibinfo{author}{Goodman, E.}, \&
  \bibinfo{author}{Banzhaf, W.} (\bibinfo{year}{2019, July}).
\newblock \bibinfo{title}{{{NSGA-Net}: Neural architecture search using
  multi-objective genetic algorithm}}.
\newblock In {\it \bibinfo{booktitle}{GECCO 2019 - Proceedings of the 2019
  Genetic and Evolutionary Computation Conference, Prague, Czech Republic,
  2019}\/}.
\newblock \bibinfo{note}{\url{https://doi.org/10.24963/ijcai.2020/659}}.
\bibitem[{Lu et~al.(2021)Lu, Whalen, Dhebar, Deb, Goodman, Banzhaf \&
  Boddeti}]{moevo1}
\bibinfo{author}{Lu, Z.}, \bibinfo{author}{Whalen, I.},
  \bibinfo{author}{Dhebar, Y.}, \bibinfo{author}{Deb, K.},
  \bibinfo{author}{Goodman, E.~D.}, \bibinfo{author}{Banzhaf, W.}, \&
  \bibinfo{author}{Boddeti, V.~N.} (\bibinfo{year}{2021}).
\newblock \bibinfo{title}{{Multiobjective Evolutionary Design of Deep
  Convolutional Neural Networks for Image Classification}}.
\newblock {\it \bibinfo{journal}{IEEE Transactions on Evolutionary
  Computation}\/},  {\it \bibinfo{volume}{25}\/}, \bibinfo{pages}{277--291}.
\newblock \bibinfo{note}{\url{https://doi.org/10.1109/TEVC.2020.3024708}}.
\bibitem[{Mart{\'\i}n et~al.(2018)Mart{\'\i}n, Lara-Cabrera, Fuentes-Hurtado,
  Naranjo \& Camacho}]{Comput2018}
\bibinfo{author}{Mart{\'\i}n, A.}, \bibinfo{author}{Lara-Cabrera, R.},
  \bibinfo{author}{Fuentes-Hurtado, F.}, \bibinfo{author}{Naranjo, V.}, \&
  \bibinfo{author}{Camacho, D.} (\bibinfo{year}{2018}).
\newblock \bibinfo{title}{{EvoDeep: a new evolutionary approach for automatic
  deep neural networks parametrisation}}.
\newblock {\it \bibinfo{journal}{Journal of Parallel and Distributed
  Computing}\/},  {\it \bibinfo{volume}{117}\/}, \bibinfo{pages}{180--191}.
\newblock \bibinfo{note}{\url{https://doi.org/10.1016/j.jpdc.2017.09.006}}.
\bibitem[{Martinez et~al.(2021)Martinez, {Del Ser}, Villar-Rodriguez, Osaba,
  Poyatos, Tabik, Molina \& Herrera}]{MARTINEZ2021161}
\bibinfo{author}{Martinez, A.~D.}, \bibinfo{author}{{Del Ser}, J.},
  \bibinfo{author}{Villar-Rodriguez, E.}, \bibinfo{author}{Osaba, E.},
  \bibinfo{author}{Poyatos, J.}, \bibinfo{author}{Tabik, S.},
  \bibinfo{author}{Molina, D.}, \& \bibinfo{author}{Herrera, F.}
  (\bibinfo{year}{2021}).
\newblock \bibinfo{title}{{Lights and shadows in Evolutionary Deep Learning:
  Taxonomy, critical methodological analysis, cases of study, learned lessons,
  recommendations and challenges}}.
\newblock {\it \bibinfo{journal}{Information Fusion}\/},  {\it
  \bibinfo{volume}{67}\/}, \bibinfo{pages}{161--194}.
\newblock \bibinfo{note}{\url{https://doi.org/10.1016/j.inffus.2020.10.014}}.
\bibitem[{Miikkulainen et~al.(2019)Miikkulainen, Liang, Meyerson, Rawal, Fink,
  Francon, Raju, Shahrzad, Navruzyan, Duffy et~al.}]{miikkulainen2019evolving}
\bibinfo{author}{Miikkulainen, R.}, \bibinfo{author}{Liang, J.},
  \bibinfo{author}{Meyerson, E.}, \bibinfo{author}{Rawal, A.},
  \bibinfo{author}{Fink, D.}, \bibinfo{author}{Francon, O.},
  \bibinfo{author}{Raju, B.}, \bibinfo{author}{Shahrzad, H.},
  \bibinfo{author}{Navruzyan, A.}, \bibinfo{author}{Duffy, N.} et~al.
  (\bibinfo{year}{2019}).
\newblock \bibinfo{title}{{Evolving deep neural networks}}.
\newblock In {\it \bibinfo{booktitle}{Artificial intelligence in the age of
  neural networks and brain computing}\/} (pp. \bibinfo{pages}{293--312}).
\newblock \bibinfo{publisher}{Academic Press}.
\newblock
  \bibinfo{note}{\url{https://doi.org/10.1016/B978-0-12-815480-9.00015-3}}.
\bibitem[{[dataset]Laurence Moroney(2019)}]{rps}
\bibinfo{author}{[dataset]Laurence Moroney} (\bibinfo{year}{2019}).
\newblock \bibinfo{title}{{Rock, Paper, Scissors Dataset}}.
\newblock \bibinfo{note}{Retrieved from
  \url{http://www.laurencemoroney.com/rock-paper-scissors-dataset/}. Accessed
  September 10, 2020}.
\bibitem[{Pan \& Yang(2010)}]{tl}
\bibinfo{author}{Pan, S.~J.}, \& \bibinfo{author}{Yang, Q.}
  (\bibinfo{year}{2010}).
\newblock \bibinfo{title}{{A Survey on Transfer Learning}}.
\newblock {\it \bibinfo{journal}{IEEE Transactions on Knowledge and Data
  Engineering}\/},  {\it \bibinfo{volume}{22}\/}, \bibinfo{pages}{1345--1359}.
\newblock \bibinfo{note}{\url{https://doi.org/10.1109/TKDE.2009.191}}.
\bibitem[{Parmar et~al.(2022)Parmar, Chouhan, Raychoudhury \& Rathore}]{owl}
\bibinfo{author}{Parmar, J.}, \bibinfo{author}{Chouhan, S.~S.},
  \bibinfo{author}{Raychoudhury, V.}, \& \bibinfo{author}{Rathore, S.~S.}
  (\bibinfo{year}{2022}).
\newblock \bibinfo{title}{{Open-World Machine Learning: Applications,
  Challenges, and Opportunities}}.
\newblock {\it \bibinfo{journal}{ACM Comput. Surv.}\/},  (pp.
  \bibinfo{pages}{1--36}).
\newblock \bibinfo{note}{\url{https://doi.org/10.1145/3561381}}.
\bibitem[{Pham et~al.(2018, July)Pham, Guan, Zoph, Le \&
  Dean}]{pham2018efficient}
\bibinfo{author}{Pham, H.}, \bibinfo{author}{Guan, M.}, \bibinfo{author}{Zoph,
  B.}, \bibinfo{author}{Le, Q.}, \& \bibinfo{author}{Dean, J.}
  (\bibinfo{year}{2018, July}).
\newblock \bibinfo{title}{{Efficient neural architecture search via parameters
  sharing}}.
\newblock In {\it \bibinfo{booktitle}{International Conference on Machine
  Learning, Stockholm, Sweden, 2018}\/}.
\newblock \bibinfo{note}{\url{http://proceedings.mlr.press/v80/pham18a.html}}.
\bibitem[{Poyatos et~al.(2023)Poyatos, Molina, Martinez, {Del Ser} \&
  Herrera}]{evoprunedeeptl}
\bibinfo{author}{Poyatos, J.}, \bibinfo{author}{Molina, D.},
  \bibinfo{author}{Martinez, A.~D.}, \bibinfo{author}{{Del Ser}, J.}, \&
  \bibinfo{author}{Herrera, F.} (\bibinfo{year}{2023}).
\newblock \bibinfo{title}{{{EvoPruneDeepTL}: An evolutionary pruning model for
  transfer learning based deep neural networks}}.
\newblock {\it \bibinfo{journal}{Neural Networks}\/},  {\it
  \bibinfo{volume}{158}\/}, \bibinfo{pages}{59--82}.
\newblock \bibinfo{note}{\url{https://doi.org/10.1016/j.neunet.2022.10.011}}.
\bibitem[{Real et~al.(2019, January)Real, Aggarwal, Huang \&
  Le}]{real2019regularized}
\bibinfo{author}{Real, E.}, \bibinfo{author}{Aggarwal, A.},
  \bibinfo{author}{Huang, Y.}, \& \bibinfo{author}{Le, Q.~V.}
  (\bibinfo{year}{2019, January}).
\newblock \bibinfo{title}{{Regularized evolution for image classifier
  architecture search}}.
\newblock In {\it \bibinfo{booktitle}{Proceedings of the AAAI Conference on
  Artificial Intelligence, Honolulu, HI, USA, 2019}\/}.
\newblock \bibinfo{note}{\url{https://doi.org/10.1609/aaai.v33i01.33014780}}.
\bibitem[{Real et~al.(2020, July)Real, Liang, So \& Le}]{real2020automl}
\bibinfo{author}{Real, E.}, \bibinfo{author}{Liang, C.}, \bibinfo{author}{So,
  D.}, \& \bibinfo{author}{Le, Q.} (\bibinfo{year}{2020, July}).
\newblock \bibinfo{title}{{AutoML-zero: evolving machine learning algorithms
  from scratch}}.
\newblock In {\it \bibinfo{booktitle}{International Conference on Machine
  Learning, 2020}\/}.
\newblock \bibinfo{note}{\url{https://doi.org/10.48550/arXiv.2003.03384}}.
\bibitem[{[dataset]Virtual Russian~Museum(2018)}]{Musemart}
\bibinfo{author}{[dataset]Virtual Russian~Museum} (\bibinfo{year}{2018}).
\newblock \bibinfo{title}{{Art Images:
  Drawing/Painting/Sculptures/Engravings}}.
\newblock \bibinfo{note}{Retrieved from
  \url{https://www.kaggle.com/thedownhill/art-images-drawings-painting-sculpture-engraving},
  Accessed September 10, 2020}.
\bibitem[{Salehi et~al.(2021)Salehi, Mirzaei, Hendrycks, Li, Rohban \&
  Sabokrou}]{salehi2021unified}
\bibinfo{author}{Salehi, M.}, \bibinfo{author}{Mirzaei, H.},
  \bibinfo{author}{Hendrycks, D.}, \bibinfo{author}{Li, Y.},
  \bibinfo{author}{Rohban, M.~H.}, \& \bibinfo{author}{Sabokrou, M.}
  (\bibinfo{year}{2021}).
\newblock \bibinfo{title}{{A unified survey on anomaly, novelty, open-set, and
  out-of-distribution detection: Solutions and future challenges}}.
\newblock {\it \bibinfo{journal}{arXiv preprint arXiv:2110.14051}\/}, .
\newblock \bibinfo{note}{\url{https://doi.org/10.48550/arXiv.2110.14051}}.
\bibitem[{Selvaraju et~al.(2017, October)Selvaraju, Cogswell, Das, Vedantam,
  Parikh \& Batra}]{gradcam}
\bibinfo{author}{Selvaraju, R.~R.}, \bibinfo{author}{Cogswell, M.},
  \bibinfo{author}{Das, A.}, \bibinfo{author}{Vedantam, R.},
  \bibinfo{author}{Parikh, D.}, \& \bibinfo{author}{Batra, D.}
  (\bibinfo{year}{2017, October}).
\newblock \bibinfo{title}{{Grad-CAM: Visual Explanations from Deep Networks via
  Gradient-Based Localization}}.
\newblock In {\it \bibinfo{booktitle}{2017 IEEE International Conference on
  Computer Vision (ICCV), Venezia, Italy, 2017}\/}.
\newblock \bibinfo{note}{\url{https://doi.org/10.1109/ICCV.2017.74}}.
\bibitem[{[dataset] Singh et~al.(2020, May)[dataset] Singh, Jain, Jain, Kayal,
  Kumawat \& Batra}]{plantdoc}
\bibinfo{author}{[dataset] Singh, D.}, \bibinfo{author}{Jain, N.},
  \bibinfo{author}{Jain, P.}, \bibinfo{author}{Kayal, P.},
  \bibinfo{author}{Kumawat, S.}, \& \bibinfo{author}{Batra, N.}
  (\bibinfo{year}{2020, May}).
\newblock \bibinfo{title}{{PlantDoc: A Dataset for Visual Plant Disease
  Detection}}.
\newblock In {\it \bibinfo{booktitle}{7th ACM IKDD CoDS and 25th COMAD,
  Hyderabad, India, 2020}\/}.
\newblock \bibinfo{note}{\url{https://doi.org/10.1145/3371158.3371196}}.
\bibitem[{Srinivas \& Babu(2015, September)}]{srinivas2015data}
\bibinfo{author}{Srinivas, S.}, \& \bibinfo{author}{Babu, R.~V.}
  (\bibinfo{year}{2015, September}).
\newblock \bibinfo{title}{{Data-free Parameter Pruning for Deep Neural
  Networks}}.
\newblock In {\it \bibinfo{booktitle}{Proceedings of the British Machine Vision
  Conference (BMVC), Swansea, UK, 2015}\/}.
\newblock
  \bibinfo{note}{\url{https://dblp.org/rec/journals/corr/SrinivasB15.bib}}.
\bibitem[{Stanley \& Miikkulainen(2002)}]{Stanley200299}
\bibinfo{author}{Stanley, K.}, \& \bibinfo{author}{Miikkulainen, R.}
  (\bibinfo{year}{2002}).
\newblock \bibinfo{title}{{Evolving neural networks through augmenting
  topologies}}.
\newblock {\it \bibinfo{journal}{Evolutionary Computation}\/},  {\it
  \bibinfo{volume}{10}\/}, \bibinfo{pages}{99--127}.
\newblock \bibinfo{note}{\url{https://doi.org/10.1162/106365602320169811}}.
\bibitem[{Suganuma et~al.(2020)Suganuma, Kobayashi, Shirakawa \&
  Nagao}]{suganuma}
\bibinfo{author}{Suganuma, M.}, \bibinfo{author}{Kobayashi, M.},
  \bibinfo{author}{Shirakawa, S.}, \& \bibinfo{author}{Nagao, T.}
  (\bibinfo{year}{2020}).
\newblock \bibinfo{title}{{Evolution of Deep Convolutional Neural Networks
  Using Cartesian Genetic Programming}}.
\newblock {\it \bibinfo{journal}{Evolutionary Computation}\/},  {\it
  \bibinfo{volume}{28}\/}, \bibinfo{pages}{141--163}.
\newblock \bibinfo{note}{\url{https://doi.org/10.1162/evco_a_00253}}.
\bibitem[{[dataset]Hafiz Tayyab~Rauf et~al.(2019)[dataset]Hafiz Tayyab~Rauf,
  Saleem, Lali, Khan, Sharif \& Bukhari}]{RAUF2019104340}
\bibinfo{author}{[dataset]Hafiz Tayyab~Rauf}, \bibinfo{author}{Saleem, B.~A.},
  \bibinfo{author}{Lali, M. I.~U.}, \bibinfo{author}{Khan, M.~A.},
  \bibinfo{author}{Sharif, M.}, \& \bibinfo{author}{Bukhari, S. A.~C.}
  (\bibinfo{year}{2019}).
\newblock \bibinfo{title}{{A citrus fruits and leaves dataset for detection and
  classification of citrus diseases through machine learning}}.
\newblock {\it \bibinfo{journal}{Data in Brief}\/},  {\it
  \bibinfo{volume}{26}\/}, \bibinfo{pages}{Article 104340}.
\newblock \bibinfo{note}{\url{https://doi.org/10.1016/j.dib.2019.104340}}.
\bibitem[{Trivedi et~al.(2018)Trivedi, Srivastava, Mishra, Shukla \&
  Tiwari}]{TRIVEDI2018525}
\bibinfo{author}{Trivedi, A.}, \bibinfo{author}{Srivastava, S.},
  \bibinfo{author}{Mishra, A.}, \bibinfo{author}{Shukla, A.}, \&
  \bibinfo{author}{Tiwari, R.} (\bibinfo{year}{2018}).
\newblock \bibinfo{title}{{Hybrid evolutionary approach for Devanagari
  handwritten numeral recognition using Convolutional Neural Network}}.
\newblock {\it \bibinfo{journal}{Procedia Computer Science}\/},  {\it
  \bibinfo{volume}{125}\/}, \bibinfo{pages}{525--532}.
\newblock \bibinfo{note}{\url{https://doi.org/10.1016/j.procs.2017.12.068}}.
\bibitem[{Wang et~al.(2019)Wang, Lin, Hu, Wang, He, Huang \&
  Chang}]{lstmpruning}
\bibinfo{author}{Wang, S.}, \bibinfo{author}{Lin, P.}, \bibinfo{author}{Hu,
  R.}, \bibinfo{author}{Wang, H.}, \bibinfo{author}{He, J.},
  \bibinfo{author}{Huang, Q.}, \& \bibinfo{author}{Chang, S.}
  (\bibinfo{year}{2019}).
\newblock \bibinfo{title}{{Acceleration of LSTM With Structured Pruning Method
  on FPGA}}.
\newblock {\it \bibinfo{journal}{IEEE Access}\/},  {\it \bibinfo{volume}{7}\/},
  \bibinfo{pages}{62930--62937}.
\newblock \bibinfo{note}{\url{10.1109/ACCESS.2019.2917312}}.
\bibitem[{Wang et~al.(2021)Wang, Liu \& Jin}]{robustintro}
\bibinfo{author}{Wang, S.}, \bibinfo{author}{Liu, J.}, \& \bibinfo{author}{Jin,
  Y.} (\bibinfo{year}{2021}).
\newblock \bibinfo{title}{{A Computationally Efficient Evolutionary Algorithm
  for Multiobjective Network Robustness Optimization}}.
\newblock {\it \bibinfo{journal}{IEEE Transactions on Evolutionary
  Computation}\/},  {\it \bibinfo{volume}{25}\/}, \bibinfo{pages}{419--432}.
\newblock \bibinfo{note}{\url{https://doi.org/10.1109/TEVC.2020.3048174}}.
\bibitem[{Wang et~al.(2020)Wang, Li, Shi, Xie \& Wang}]{WANG2020247}
\bibinfo{author}{Wang, Z.}, \bibinfo{author}{Li, F.}, \bibinfo{author}{Shi,
  G.}, \bibinfo{author}{Xie, X.}, \& \bibinfo{author}{Wang, F.}
  (\bibinfo{year}{2020}).
\newblock \bibinfo{title}{{Network pruning using sparse learning and genetic
  algorithm}}.
\newblock {\it \bibinfo{journal}{Neurocomputing}\/},  {\it
  \bibinfo{volume}{404}\/}, \bibinfo{pages}{247--256}.
\newblock \bibinfo{note}{\url{https://doi.org/10.1016/j.neucom.2020.03.082}}.
\bibitem[{Wei et~al.(2022)Wei, Lee, Hu \& Chen}]{moodnas}
\bibinfo{author}{Wei, H.}, \bibinfo{author}{Lee, F.}, \bibinfo{author}{Hu, C.},
  \& \bibinfo{author}{Chen, Q.} (\bibinfo{year}{2022}).
\newblock \bibinfo{title}{{MOO-DNAS: Efficient Neural Network Design via
  Differentiable Architecture Search Based on Multi-Objective Optimization}}.
\newblock {\it \bibinfo{journal}{IEEE Access}\/},  {\it
  \bibinfo{volume}{10}\/}, \bibinfo{pages}{14195--14207}.
\newblock \bibinfo{note}{\url{https://doi.org/10.1109/ACCESS.2022.3148323}}.
\bibitem[{Yang et~al.(2021)Yang, Zhou, Li \& Liu}]{yang2021generalized}
\bibinfo{author}{Yang, J.}, \bibinfo{author}{Zhou, K.}, \bibinfo{author}{Li,
  Y.}, \& \bibinfo{author}{Liu, Z.} (\bibinfo{year}{2021}).
\newblock \bibinfo{title}{{Generalized out-of-distribution detection: A
  survey}}.
\newblock {\it \bibinfo{journal}{arXiv preprint arXiv:2110.11334}\/}, .
\newblock \bibinfo{note}{\url{https://doi.org/10.48550/arxiv.2110.11334}}.
\bibitem[{Yang et~al.(2020, June)Yang, Wang, Chen, Shi, Xu, Xu, Tian \&
  Xu}]{supernet}
\bibinfo{author}{Yang, Z.}, \bibinfo{author}{Wang, Y.}, \bibinfo{author}{Chen,
  X.}, \bibinfo{author}{Shi, B.}, \bibinfo{author}{Xu, C.},
  \bibinfo{author}{Xu, C.}, \bibinfo{author}{Tian, Q.}, \& \bibinfo{author}{Xu,
  C.} (\bibinfo{year}{2020, June}).
\newblock \bibinfo{title}{{CARS: Continuous Evolution for Efficient Neural
  Architecture Search}}.
\newblock In {\it \bibinfo{booktitle}{2020 IEEE/CVF Conference on Computer
  Vision and Pattern Recognition (CVPR), Virtual, June, 2020}\/}.
\newblock \bibinfo{note}{\url{https://doi.org/10.1109/CVPR42600.2020.00190}}.
\bibitem[{Zhou(2022)}]{owlml}
\bibinfo{author}{Zhou, Z.-H.} (\bibinfo{year}{2022}).
\newblock \bibinfo{title}{{Open-environment machine learning}}.
\newblock {\it \bibinfo{journal}{National Science Review}\/},  {\it
  \bibinfo{volume}{9}\/}, \bibinfo{pages}{1--11}.
\newblock \bibinfo{note}{\url{https://doi.org/10.1093/nsr/nwac123}}.
\bibitem[{Zoph \& Le(2016)}]{nasrl}
\bibinfo{author}{Zoph, B.}, \& \bibinfo{author}{Le, Q.~V.}
  (\bibinfo{year}{2016}).
\newblock \bibinfo{title}{{Neural Architecture Search with Reinforcement
  Learning}}.
\newblock {\it \bibinfo{journal}{arXiv preprint arXiv:1611.01578}\/}, .
\newblock \bibinfo{note}{\url{https://arxiv.org/abs/1611.01578}}.

\end{thebibliography}

\end{document}